\documentclass[10pt,twocolumn,letterpaper]{article}

\usepackage[pagenumbers]{wacv} 

\definecolor{wacvblue}{rgb}{0.21,0.49,0.74}
\usepackage[pagebackref,breaklinks,colorlinks,allcolors=wacvblue]{hyperref}
\usepackage{multirow}
\usepackage[most]{tcolorbox} 
\usepackage{tabularray}
\usepackage{siunitx}
\usepackage{subcaption}
\usepackage{microtype}

\usepackage[dvipsnames]{xcolor}
\definecolor{blue}{HTML}{5F80B4}
\definecolor{green}{HTML}{6F9759}
\definecolor{orange}{HTML}{B88700}

\newtcolorbox{inlinenote}[2][]{%
  enhanced,
  breakable,
  colback=#2!10,
  colframe=#2!60!black,
  left=1pt,right=1pt,top=1pt,bottom=1pt,
  boxrule=0.6pt,arc=2pt,
}

\usepackage[table]{xcolor}
\definecolor{best}{rgb}{0.70, 0.88, 0.70}   
\definecolor{second}{rgb}{0.85, 0.95, 0.85} 

\def\wacvPaperID{195} 
\def\confName{WACV}
\def\confYear{2027}

\title{Expand-and-Prune: Test-Time Candidate-Space Adaptation for\\End-to-End Autonomous Driving}

\author{
Fabian Schmidt\textsuperscript{1,2} \qquad
Karol Fedurko\textsuperscript{1} \qquad
Markus Enzweiler\textsuperscript{1} \qquad
Abhinav Valada\textsuperscript{2}\\[10pt]
\textsuperscript{1}Esslingen University of Applied Sciences \qquad
\textsuperscript{2}University of Freiburg\\
}

\begin{document}
\maketitle

\begin{abstract}
Candidate-based end-to-end driving planners reason over a finite set of trajectory hypotheses. However, for sparse planners, a limited candidate set may provide insufficient support for scene-relevant maneuvers, while fixed downstream candidate budgets do not account for prediction uncertainty. 
To address these challenges, we introduce \textit{Expand-and-Prune}, a test-time framework that jointly adapts candidate support and capacity without updating learned parameters. 
The framework expands the native candidate set using semantic driving intent and temporally propagated predictions and then uses uncertainty in intermediate trajectory scores to adaptively prune the resulting candidates.
Across four heterogeneous candidate-based planners and four benchmarks, \textit{Expand-and-Prune} consistently improves planning performance.
It achieves state-of-the-art results on LangAuto, increasing the Driving Score from $68.2$ to $75.9$, while providing further improvement on Bench2Drive, Fail2Drive, and NAVSIMv2, with particularly strong improvements in out-of-distribution scenarios.
Moreover, in its full instantiation, $72.5\%$ of selected trajectories originate from the newly introduced intent-guided or temporal candidates, confirming that candidate-space expansion provides useful hypotheses beyond those generated natively by the planner.
The code is available at \url{https://github.com/iis-esslingen/Expand-and-Prune}.
\end{abstract}

\section{Introduction}
\label{sec:intro}

End-to-end planning has become a prominent paradigm for autonomous driving, directly mapping sensor observations and contextual information to future ego motion \cite{chitta2022transfuser,hu2023uniad,jiang2023vad,schmidt2025graphpilot}.
Rather than predicting only a single trajectory, many recent planners capture multimodal driving behavior through a finite set of motion hypotheses. 
These candidates may originate from offline precomputed trajectory vocabularies or clustered anchors \cite{li2024hydramdp,li2025gtrs,yao2026drivesuprim,sun2026sparsedrivev2}, or be generated and refined online through learned queries, diffusion, or related generative mechanisms \cite{kirby2026drivor,feng2026rap,liao2025diffusiondrive,zou2025diffusiondrivev2,schmidt2026laddrive}.
The planner subsequently scores or refines these hypotheses before selecting a trajectory for execution \cite{li2025gtrs,yao2026drivesuprim,sun2026sparsedrivev2}.

\begin{figure}[t!]
    \centering
    \includegraphics[width=1.0\linewidth]
        {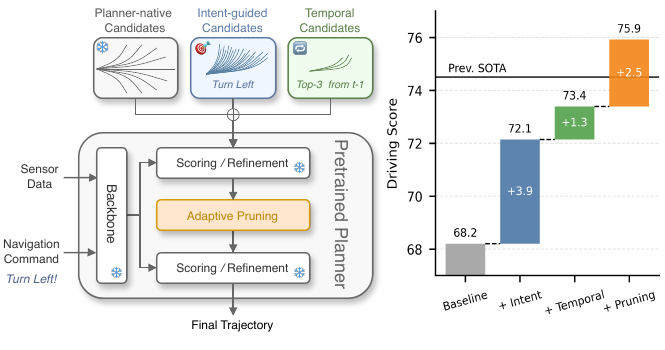}
    \vspace{-0.5cm}
    \caption{
    Concept and performance of \emph{Expand-and-Prune}.
    \textcolor{blue}{Intent-guided} and \textcolor{green}{temporal} candidates expand the planner-native candidate set, while
    \textcolor{orange}{adaptive pruning} regulates downstream capacity. On LangAuto, \emph{Expand-and-Prune} improves LAD-Drive from $68.2$ to $75.9$ Driving Score, achieving SOTA performance.
    }
    \label{fig:cover_overview}
    \vspace{-0.4cm}
\end{figure}

Despite advances in candidate generation, how motion hypotheses are constructed and how many are propagated downstream remains largely prescribed by the planner's inference design. 
First, for sparse anchor- or proposal-based planners \cite{liao2025diffusiondrive, liu2026bridgedrive}, a limited number of hypotheses must cover a broad motion space and may therefore provide only limited support for the maneuver currently favored by the scene. 
For example, a generic candidate set may contain only a few hypotheses matching a sharp turn even when the current semantic belief strongly favors that maneuver. 
Second, a fixed downstream candidate budget, often implemented through static Top-$K$ selection, treats confident and ambiguous scenes alike, propagating the same number of hypotheses regardless of how many plausible alternatives remain \cite{yao2026drivesuprim, sun2026sparsedrivev2}. 
Such limitations may become particularly consequential over longer closed-loop horizons and under distribution shift, where the planner may encounter scene configurations and maneuvers that are poorly represented by its training distribution and native candidate set.



Learned scene-adaptive proposal mechanisms \cite{yao2026had,yin2026diffrefiner} can improve candidate generation, but require planner-specific training. Other methods operate at test time \cite{sima2025centaur,xu2026toad}, either updating model parameters or optimizing trajectories using the planner's scores as an objective. In contrast, we keep the pretrained planner entirely fixed and adapt its candidate space at inference time, leaving both the model parameters and the scoring procedure unchanged. Thus, we enable adaptation of pretrained planners without specific retraining or modification of their native scoring pipeline.


Building on this principle, we introduce \emph{Expand-and-Prune}, which dynamically reshapes the planner's candidate space through complementary expansion and pruning, cf. Fig.~\ref{fig:cover_overview}.
Intent-guided candidate expansion and entropy-adaptive temporal memory augment the native candidate set with semantically and temporally relevant hypotheses. 
Uncertainty-aware adaptive pruning regulates downstream candidate capacity. 
Expansion therefore adapts \emph{where} candidate support is allocated, while pruning adapts \emph{how many} hypotheses receive further computation.
Both mechanisms operate at existing candidate-processing boundaries, enabling their integration into different candidate-based architectures. 

We evaluate \emph{Expand-and-Prune} using LAD-Drive~\cite{schmidt2026laddrive} on LangAuto~\cite{shao2024lmdrive} as our primary testbed. \emph{Expand-and-Prune} achieves state-of-the-art performance, increasing mean Driving Score from $68.2$ to $75.9$, with particularly strong improvements on longer routes. 
We further evaluate the transferability of our framework across three additional heterogeneous candidate-based planners.
For each architecture, we instantiate the components that are compatible with its native candidate representation and processing pipeline, using the established evaluation settings: BridgeDrive~\cite{liu2026bridgedrive} on Bench2Drive~\cite{jia2024bench2drive}, and DriveSuprim~\cite{yao2026drivesuprim} and SparseDriveV2~\cite{sun2026sparsedrivev2} on NAVSIMv2~\cite{Cao2025navsimv2}. 
Across these planners and benchmarks, the proposed adaptations consistently improve planning performance. 
We additionally evaluate our candidate-space-adapted BridgeDrive on Fail2Drive~\cite{gerstenecker2026fail2drive}, where the gains become particularly pronounced in out-of-distribution scenarios. 
Candidate-provenance analysis on LAD-Drive further reveals that $72.5\%$ of selected trajectories originate from the added intent-guided or temporal candidates, indicating that the additional hypotheses frequently influence the planner's final decision.\looseness=-1


Our main contribution is \emph{Expand-and-Prune}, a test-time framework for adapting the candidate support and downstream capacity of pretrained candidate-based driving planners without updating their learned parameters or modifying their scoring procedure. 
We demonstrate the generality of this principle across sparse diffusion and dense selection-based planners, complemented by analyses of candidate provenance, adaptive capacity, and computational efficiency. 


\section{Related Work}
\label{sec:related_work}

\textbf{Candidate-based end-to-end planning}. Many end-to-end driving systems directly predict a single ego trajectory or waypoint sequence from sensor observations \cite{chitta2022transfuser,wu2022tcp}, without explicitly representing multiple plausible future behaviors.
More recent approaches model several trajectory hypotheses using offline trajectory vocabularies, clustered anchors, or learned query-based proposal generators \cite{li2024hydramdp,li2025hydranext,li2025gtrs,li2025ztrs,jiang2026vadv2,guo2025ipad,sun2026sparsedrivev2,yao2026drivesuprim,kirby2026drivor,feng2026rap}.
For instance, DrivoR \cite{kirby2026drivor} decodes candidates from learned trajectory queries and evaluates them using a separate scoring decoder.
Another line of work progressively refines trajectory hypotheses using diffusion, flow matching, or diffusion bridges \cite{liao2025diffusiondrive,zou2025diffusiondrivev2,xing2025goalflow,liu2026bridgedrive,schmidt2026laddrive,yin2026diffrefiner,song2026diver}.
HAD \cite{yao2026had} generates structured local variations around selected intention anchors, while DiffRefiner \cite{yin2026diffrefiner} adjusts predefined anchors using a learned proposal decoder before applying scene-conditioned diffusion.
Although these methods generate scene-aware hypotheses, their candidate generation and refinement are tightly coupled to model-specific architectures trained jointly with the planner. 
In contrast, our framework modifies the candidate collection of an already trained planner directly at inference time.

\noindent \textbf{Test-time planning adaptation}. Deployment-time methods improve planning through online parameter updates, iterative trajectory search, score adaptation, or output-level safety mechanisms.
Centaur \cite{sima2025centaur} updates planner parameters using gradient steps driven by a cluster-level entropy objective.
TOAD \cite{xu2026toad} keeps the planner frozen and treats its scoring module as a reward for iterative Cross-Entropy Method optimization in continuous control space.
Diffusion-ES \cite{yang2024diffusiones} combines a trained diffusion model with gradient-free evolutionary search, iteratively selecting and mutating trajectory populations according to a black-box reward.
DrivoR \cite{kirby2026drivor} adapts driving behavior by reweighting predicted trajectory sub-scores at inference.
UniAD \cite{hu2023uniad} and SafetyNet \cite{vitelli2022safetynet} employ occupancy-aware trajectory refinement and rule-based safety validation with fallback behavior, respectively.

Our work considers a complementary paradigm: \emph{test-time candidate-space adaptation}. 
Rather than updating planner parameters, modifying only the final scoring objective, or introducing an external iterative trajectory optimizer, we adapt the candidate collection within the planner's native inference pipeline. 
Candidate support is expanded using semantic intent and temporal memory, while uncertainty in intermediate trajectory scores adapts the downstream candidate capacity. 
All learned planner parameters remain unchanged.

\begin{figure*}[t!]
    \centering
    \includegraphics[width=\textwidth]{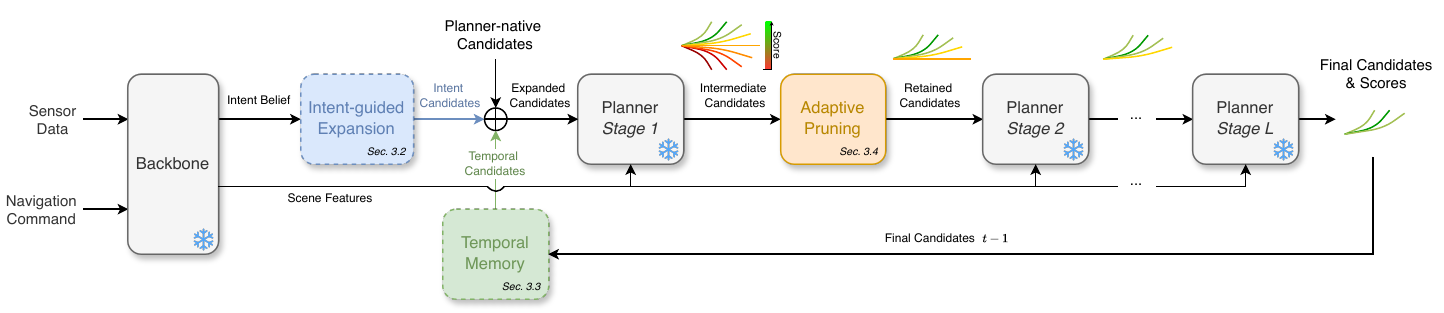}
    \caption{Overview of \emph{Expand-and-Prune}. Planner-native candidates are augmented with \textcolor{blue}{intent-guided} and \textcolor{green}{temporal} candidates, processed by the frozen planner, and \textcolor{orange}{adaptively pruned} based on intermediate scores before further refinement. The \emph{backbone} denotes the upstream perception and context-encoding components, including vision encoders and, where applicable, language-based modules, while \emph{planner stages} denote candidate-processing modules such as scoring or trajectory refinement. Dashed boxes indicate optional modules.}
    \label{fig:method_overview}

    \vspace{-0.75em}
\end{figure*}

\section{Method}
\label{sec:method}

\subsection{Problem Formulation and Overview}
\label{sec:method_overview}

Candidate-based end-to-end planners represent multiple plausible future driving behaviors through a finite collection of motion hypotheses. 
At planning step $t$, let $\mathcal{C}_{t}^{\mathrm{base}}=\{\mathbf{c}_{t,i}\}_{i=1}^{N_t}$ denote the planner's native candidate collection, where $\mathbf{c}_{t,i}$ is the $i$-th motion hypothesis and $N_t$ denotes the number of candidates.
Depending on the underlying architecture, these hypotheses may correspond to trajectory anchors, coarse motion proposals, path hypotheses, or hierarchical path--velocity representations. 
Conditioned on the current scene, the planner processes and potentially refines the candidate trajectories through one or more stages. 
During this process, the hypotheses are assigned scores that reflect the planner's relative preference, and the highest-scoring trajectory is typically selected for execution. 
We index the processing stages by $\ell\in\{1,\ldots,L\}$. 
If candidate scores $s_t$ are available at stage $\ell$, we denote them by $\mathbf{s}_{t}^{(\ell)}=[s_{t,1}^{(\ell)},\ldots,s_{t,N_t^{(\ell)}}^{(\ell)}]$, where $N_t^{(\ell)}$ denotes the number of candidates at that stage.

The mechanisms used to construct the native candidate collection $\mathcal{C}_{t}^{\mathrm{base}}$, which contains $N_t$ motion hypotheses, as well as the candidate budgets used in subsequent processing stages, are generally determined by the pretrained architecture.
In anchor-based planners, the native candidate collection is often drawn from a global prototype bank that is reused across scenes and planning steps, whereas other architectures generate scene-dependent candidates via learned queries or proposal generators.
For sparse anchor- or proposal-based planners, a limited candidate set may provide insufficient support for hypotheses that are particularly relevant to the current semantic intent or recent temporal context. 
Moreover, fixed downstream candidate budgets do not account for the uncertainty of the planner's current prediction.\looseness=-1

\noindent \textbf{Primary planner.}
We use LAD-Drive~\cite{schmidt2026laddrive} as our primary testbed as its architecture exposes two complementary signals that are particularly well-suited to candidate-space adaptation. 
First, its action decoder predicts a probabilistic belief over six meta-actions: \textit{Lane Follow}, \textit{Straight}, \textit{Left}, \textit{Right}, \textit{Lane Change Left}, and \textit{Lane Change Right}, providing an explicit representation of semantic uncertainty.
Second, its diffusion decoder initializes planning from $20$ trajectory anchors and refines them over two denoising steps while predicting candidate-level trajectory confidence scores. 
We use the action belief to guide candidate expansion and the intermediate trajectory scores to adapt the number of hypotheses retained for further refinement. 
All learned components and parameters remain unchanged.

As illustrated in Fig.~\ref{fig:method_overview}, at inference time, \emph{Expand-and-Prune} augments the native candidate collection with intent-guided and temporally propagated hypotheses:
\begin{equation}
    \mathcal{C}_{t}^{(0)}
    =
    \mathcal{C}_{t}^{\mathrm{base}}
    \uplus
    \mathcal{C}_{t}^{\mathrm{int}}
    \uplus
    \mathcal{C}_{t}^{\mathrm{temp}},
    \label{eq:expanded_candidate_set}
\end{equation}
where $\uplus$ denotes multiset union, $\mathcal{C}_{t}^{\mathrm{int}}$ consists of intent-guided hypotheses, and $\mathcal{C}_{t}^{\mathrm{temp}}$ contains temporally propagated hypotheses from the preceding frame, respectively.
Once meaningful intermediate scores become available, the expanded collection is adaptively pruned according to their uncertainty, and only the retained hypotheses are propagated through the remaining planner stages. 

\subsection{Intent-Guided Candidate Expansion}
\label{sec:intent_expansion}

Many candidate-based planners derive their trajectory hypotheses from global motion priors learned or constructed from the training distribution \cite{liao2025diffusiondrive}. 
While such candidates provide broad geometric coverage, their density is not necessarily aligned with the semantic driving intent of the current scene.
Therefore, we construct a set of trajectory prototype banks $\mathbb{B}=\{\mathcal{B}_a\}_{a\in\mathcal{A}}$, where $\mathcal{A}$ denotes a discrete semantic intent space and each $\mathcal{B}_a$ contains representative trajectories associated with intent $a$. 
In our primary LAD-Drive instantiation, these intents correspond to its six meta-actions. 
Hence, we refer to the resulting prototype collections as \emph{action banks}.
At inference time, a semantic belief over $\mathcal{A}$ determines which regions of this motion space are emphasized when expanding the planner's native candidate collection.

\begin{figure}[t!]
    \centering
    \includegraphics[width=1.0\linewidth]
        {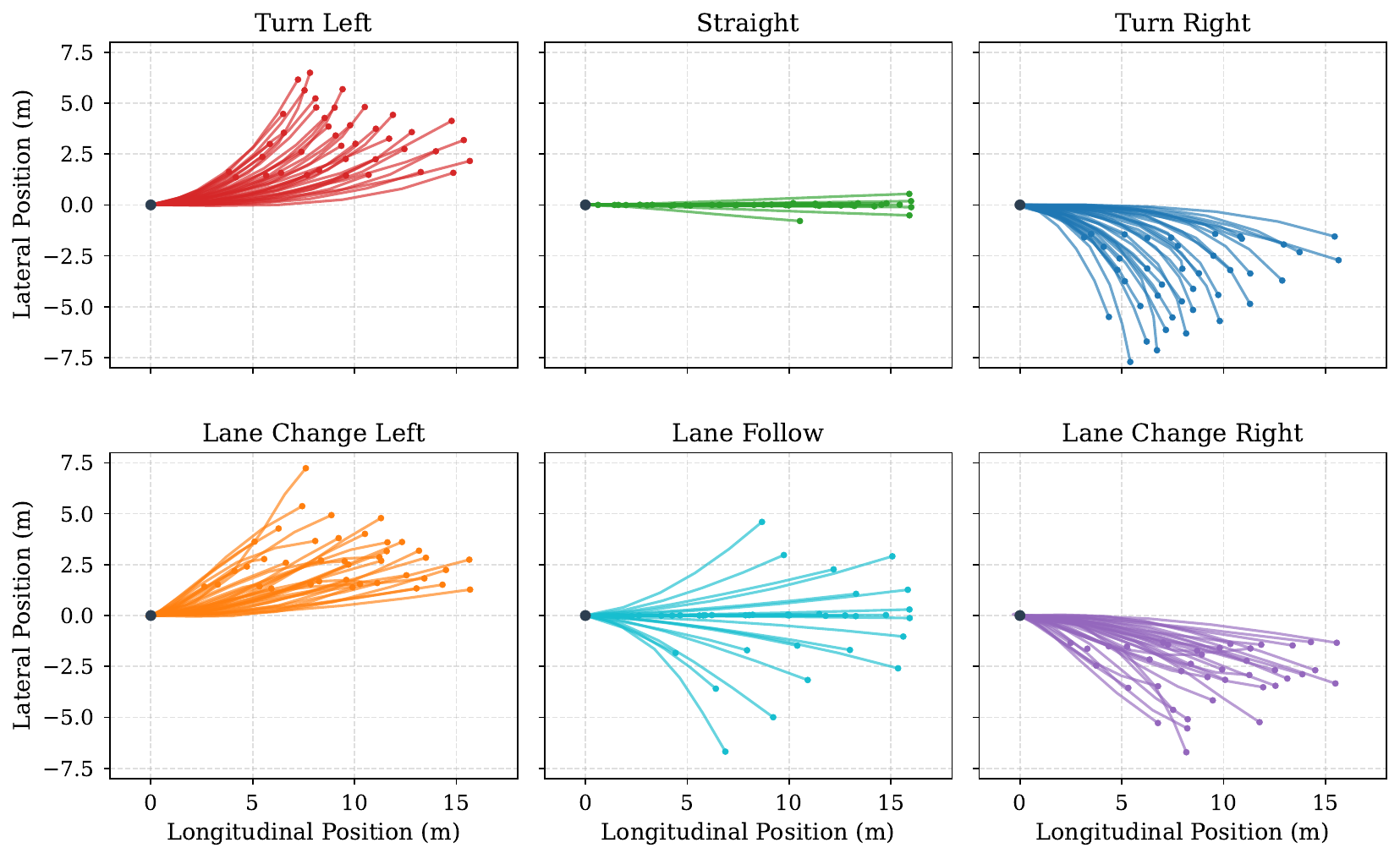}
    \caption{
        LAD-Drive action-specific trajectory prototype banks.
    }
    \label{fig:action_banks}

    \vspace{-0.75em}
    
\end{figure}

\noindent \textbf{Action-bank construction.}
Given an expert trajectory dataset $\mathcal{D}_{\mathrm{traj}}=\{\boldsymbol{\tau}_n\}_{n=1}^{N_{\mathrm{data}}}$, we assign each trajectory $\tau$ to an action category using a deterministic labeling function $\phi:\mathbb{R}^{T\times d}\rightarrow\mathcal{A}$.
The labeling heuristics characterize the geometric behavior of the trajectory using quantities such as lateral displacement and heading evolution. 
We partition the expert trajectories by their assigned actions and independently apply $k$-means clustering within each subset, retaining $M_a$ trajectory prototypes for each action $a\in\mathcal{A}$. 
The resulting action-specific candidate banks, as shown in Fig.~\ref{fig:action_banks}, are constructed once offline.

\noindent \textbf{Intent-conditioned candidate allocation.}
At planning step $t$, let $\mathbf{q}_t=[q_{t,1},\ldots,q_{t,|\mathcal{A}|}]$ denote a semantic belief over the intent space $\mathcal{A}$, where $q_{t,a}\in[0,1]$ is the probability assigned to intent $a$ and $\sum_{a\in\mathcal{A}} q_{t,a}=1$.
Given a total intent-expansion budget $N_{\mathrm{int}}$, the belief mass determines how densely the corresponding action banks are sampled.
Specifically, approximately $n_{t,a}\approx N_{\mathrm{int}}\cdot q_{t,a}$ candidates are allocated to intent $a$, with the resulting integer counts chosen such that $\sum_{a\in\mathcal{A}} n_{t,a}=N_{\mathrm{int}}$.
The intent-guided candidate set is then obtained as
\begin{equation}
    \mathcal{C}_{t}^{\mathrm{int}}
    =
    \biguplus_{a\in\mathcal{A}}
    \operatorname{Sample}
    \left(
        \mathcal{B}_a,
        n_{t,a}
    \right),
    \label{eq:intent_candidate_set}
\end{equation}
where prototypes are sampled randomly from the corresponding banks.
Consequently, intents with high belief mass receive a denser set of geometric hypotheses, leaving alternative maneuvers represented according to their probability.

\noindent \textbf{Probabilistic intent guidance.}
In our primary instantiation, the semantic belief $\mathbf{q}_t$ is provided directly by LAD-Drive's learned probabilistic action belief.
The probabilistic belief is particularly useful for candidate-space expansion, as ambiguous scenes need not be reduced to a single winner-takes-all maneuver. 
Instead, the expansion budget can be distributed across several plausible intents according to their predicted likelihoods.
The semantic belief can be derived from other planner inputs, such as categorical navigation commands or continuous goal points, with the latter mapped to the discrete intent space based on their relative geometry.

\subsection{Entropy-Adaptive Temporal Memory}
\label{sec:temporal_memory}

Predicted ego trajectories are strongly correlated across consecutive planning frames.
We exploit this temporal structure by reusing high-confidence predictions from the previous frame as additional candidate hypotheses. 
Rather than propagating a fixed number of trajectories, we adapt the temporal-memory capacity to the uncertainty of the preceding prediction, retaining more alternatives in ambiguous situations and fewer temporal hypotheses when the planner is confident.

Given the terminal trajectory scores $\mathbf{s}_{t-1}\in\mathbb{R}^{N_{t-1}}$ over the $N_{t-1}$ candidates from the preceding planning step, we obtain the confidence distribution $\mathbf{p}_{t-1}=\operatorname{softmax}(\mathbf{s}_{t-1}/\tau)$, where temperature $\tau$ controls its sharpness. 
We quantify the prediction uncertainty $h_{t-1}$ using normalized entropy,
\begin{equation}
    h_{t-1}
    =
    -\frac{
        \sum_i p_{t-1,i}\log p_{t-1,i}
    }{\log N_{t-1}} .
    \label{eq:memory_entropy}
\end{equation}
The number of trajectories propagated to the current frame is obtained via linear interpolation between the minimum and maximum memory capacities $K_{\min}^{\mathrm{mem}}$ and $K_{\max}^{\mathrm{mem}}$:
\begin{equation}
    K^{\mathrm{mem}}_t
    =
    \left\lfloor
        K^{\mathrm{mem}}_{\min}
        +
        h_{t-1}
        \left(
            K^{\mathrm{mem}}_{\max}
            -
            K^{\mathrm{mem}}_{\min}
        \right)
    \right\rceil .
    \label{eq:memory_budget}
\end{equation}
Thus, confident predictions approach the minimum memory budget, whereas high-entropy predictions preserve a broader set of alternatives. 
We retain the $K^{\mathrm{mem}}_t$ highest-scoring trajectories from frame $t-1$.

The retained trajectories are expressed in the previous ego frame. Let $\mathbf{T}_t^w \in \mathrm{SE}(2)$ map ego coordinates to world coordinates at time $t$. 
To compensate for ego translation and rotation, we apply the relative transformation $\mathbf{T}_{t\leftarrow t-1} = (\mathbf{T}_t^w)^{-1}\mathbf{T}_{t-1}^w$ to each waypoint in homogeneous coordinates, without temporal resampling. 
The resulting trajectories form $\mathcal{C}^{\mathrm{temp}}_t$, providing a scene-adaptive prior for the current planning step.

\subsection{Confidence-Aware Adaptive Pruning}
\label{sec:adaptive_pruning}

Candidate expansion broadens trajectory-space coverage, but also increases the number of hypotheses entering downstream refinement stages. 
Propagating all expanded candidates through the complete planner therefore introduces additional computational cost, even when the intermediate prediction is already concentrated on a small subset of plausible trajectories. 
A fixed top-$K$ strategy only partially addresses this issue, as it retains the same number of candidates irrespective of prediction uncertainty.
However, the required candidate capacity depends on the ambiguity of the current scene: when confidence is concentrated on a few hypotheses, aggressive pruning is sufficient, whereas uncertain predictions benefit from preserving a broader set of alternatives. 
Thus, we adapt the pruning budget to the uncertainty of the intermediate trajectory-score distribution.

Let $\ell_p$ denote the earliest planner stage that provides meaningful candidate scores, and let $N_t^{(\ell_p)} = |\mathcal{C}_t^{(\ell_p)}|$ denote the number of candidates available at that stage.
Given the corresponding scores $\mathbf{s}_{t}^{(\ell_p)}=[s_{t,1}^{(\ell_p)},\ldots,s_{t,N_t^{(\ell_p)}}^{(\ell_p)}]$, we obtain the temperature-scaled confidence distribution $\mathbf{p}_{t}^{(\ell_p)}=\operatorname{softmax}(\mathbf{s}_{t}^{(\ell_p)}/\tau)$ and compute the normalized entropy $h_t^{(\ell_p)}\in[0,1]$ analogously to Eq.~\eqref{eq:memory_entropy}, using $N_t^{(\ell_p)}$ for normalization.
The number of candidates retained for refinement is obtained via linear interpolation between the minimum and maximum pruning budgets $K_{\min}^{\mathrm{prune}}$ and $K_{\max}^{\mathrm{prune}}$:
\begin{equation}
    K^{\mathrm{prune}}_t
    =
    \left\lfloor
        K^{\mathrm{prune}}_{\min}
        +
        h_t^{(\ell_p)}
        \left(
            K^{\mathrm{prune}}_{\max}
            -
            K^{\mathrm{prune}}_{\min}
        \right)
    \right\rceil .
    \label{eq:pruning_budget}
\end{equation}
Hence, low-entropy predictions approach the minimum candidate budget, while high-entropy predictions retain additional alternatives. 
We preserve the $K_t^{\mathrm{prune}}$ highest-scoring hypotheses and propagate only these candidates through the remaining planner stages, thereby reducing downstream computation while concentrating refinement on the most promising alternatives.

\subsection{Integration into Candidate-Based Planners}
\label{sec:planner_integration}

\emph{Expand-and-Prune} modifies candidate collections only at existing candidate-processing boundaries and leaves the learned planner parameters unchanged. 
In multi-step diffusion planners, expansion precedes denoising, and adaptive pruning can be applied between denoising stages. 
In coarse-to-fine or hierarchical planners, the same principle can be instantiated at native proposal-selection or refinement boundaries. 
The specific adaptations for the evaluated architectures are detailed in Sec.~\ref{sec:experimental_setup}.

\section{Experiments}
\label{sec:experiments}


\subsection{Experimental Setup}
\label{sec:experimental_setup}

\noindent \textbf{Benchmarks and metrics.}
We evaluate \emph{Expand-and-Prune} across closed-loop simulation and real-world planning benchmarks. 
For closed-loop evaluation, we use LangAuto~\cite{shao2024lmdrive}, Bench2Drive~\cite{jia2024bench2drive}, and Fail2Drive~\cite{gerstenecker2026fail2drive}, all built on the CARLA simulator~\cite{dosovitskiy2017carla}. LangAuto evaluates driving across eight towns and three route-length categories: \textit{Tiny} ($<150\,\mathrm{m}$), \textit{Short} ($150$--$500\,\mathrm{m}$), and \textit{Long} ($>500\,\mathrm{m}$).
We report Route Completion (RC), the Infraction Score (IS), and the resulting Driving Score (DS), defined as their product. 
Bench2Drive evaluates 220 short routes covering 44 interactive scenarios across 15 towns. 
In addition to DS, we report Success Rate (SR), i.e., the fraction of routes completed with a perfect driving score. 
Fail2Drive complements this standard closed-loop evaluation with failure-oriented scenarios in both \emph{base} and \emph{generalization} settings, targeting robustness to out-of-distribution conditions.
In line with our evaluation protocol, we also report the harmonic mean (HM) of DS and SR. 
For complementary evaluation on real-world driving data, we use NAVSIMv2~\cite{Cao2025navsimv2}, which extends the original NAVSIM~\cite{dauner2024navsim} evaluation with additional criteria for driving-direction compliance, traffic-light compliance, lane keeping, and extended comfort. 
Performance is summarized by the Extended PDM Score (EPDMS), which aggregates these criteria with measures of collision avoidance, drivable-area compliance, progress, time-to-collision, and comfort. 
Further details on the benchmarks and metrics are provided in the supplementary material.

\begin{table}[t]
\centering

\caption{
Planner-specific instantiations of \emph{Expand-and-Prune}.
}

\label{tab:planner_adaptation}
\footnotesize
\setlength{\tabcolsep}{4.0pt}
\begin{tabular}{lcccc}
\toprule
Planner
& Method
& Intent
& Temporal
& Pruning \\
\midrule
LAD-Drive~\cite{schmidt2026laddrive}
& Sparse diffusion
& \checkmark
& \checkmark
& \checkmark \\

BridgeDrive~\cite{liu2026bridgedrive}
& Sparse diffusion
& --
& \checkmark
& \checkmark \\

DriveSuprim~\cite{yao2026drivesuprim}
& Dense selection
& --
& --
& \checkmark \\

SparseDriveV2~\cite{sun2026sparsedrivev2}
& Dense selection
& --
& --
& \checkmark \\
\bottomrule
\end{tabular}

\vspace{-0.75em}

\end{table}

\begin{table*}[ht!]
  \caption{Closed-loop performance on LangAuto. Best results are highlighted in bold.}
  \label{tab:prompt_config}
  \footnotesize
  \centering
  \begin{tabular}{l c ccc ccc ccc | ccc}
    \toprule
    \multicolumn{1}{c}{Method} & Venue & 
      \multicolumn{3}{c}{Tiny} & 
      \multicolumn{3}{c}{Short} & 
      \multicolumn{3}{c}{Long} & 
      \multicolumn{3}{|c}{Mean} \\
    \cmidrule(lr){3-5}\cmidrule(lr){6-8}\cmidrule(lr){9-11}\cmidrule(l){12-14}
    & & DS $\uparrow$ & RC $\uparrow$ & IS $\uparrow$ & DS $\uparrow$ & RC $\uparrow$ & IS $\uparrow$ & DS $\uparrow$ & RC $\uparrow$ & IS $\uparrow$ & DS $\uparrow$ & RC $\uparrow$ & IS $\uparrow$ \\
    \midrule
    DSDrive~\cite{liu2025dsdrive}           & arXiv'25 & 60.7 & 72.5 & 0.84 & 62.1 & 76.2 & 0.81 & 29.6 & 39.3 & 0.77 & 50.8 & 62.7 & 0.81 \\
    LMDrive~\cite{shao2024lmdrive}          & CVPR'24  & 66.5 & 77.9 & 0.85 & 50.6 & 60.0 & 0.84 & 36.2 & 46.6 & 0.81 & 51.1 & 61.5 & 0.83 \\
    AD-H~\cite{zhang2024adh}                & arXiv'24 & 77.5 & 85.1 & 0.91 & 56.1 & 68.0 & 0.78 & 44.0 & 53.2 & 0.83 & 59.2 & 68.8 & 0.84 \\
    BEVDriver~\cite{winter2025bevdriver}    & IROS'25  & 70.2 & 81.3 & 0.87 & 66.7 & 77.8 & 0.87 & 48.9 & 59.7 & 0.82 & 61.9 & 72.9 & 0.85 \\
    SToRM~\cite{kim2026storm}               & ICRA'26 & 78.8 & 86.9 & 0.92 & 64.5 & 74.7 & 0.88 & 44.2 & 56.8 & 0.82 & 62.5 & 72.8 & 0.87 \\
    VLDrive~\cite{zhang2025vldrive}         & ICCV'25  & 81.9 & 85.5 & 0.94 & 67.4 & 78.1 & 0.85 & 43.8 & 54.5 & 0.84 & 64.4 & 72.7 & 0.88 \\
    AdaDrive~\cite{zhang2025adadrive}       & ICCV'25  & 80.9 & 87.6 & 0.90 & 70.6 & 85.3 & 0.81 & 42.9 & 53.4 & 0.82 & 64.8 & 75.4 & 0.84 \\
    SAMoE-VLA ~\cite{you2026samoe}          & arXiv'26 & 79.5 & 86.4 & 0.92 & 69.5 & 80.9 & 0.86 & 51.4 & 63.5 & 0.81 & 66.8 & 76.9 & 0.86 \\
    LMGenDrive~\cite{shao2026lmgendrive}    & ECCV'26 & 84.1 & \textbf{92.5} & 0.92 & 77.1 & \textbf{87.9} & 0.88 & \textbf{62.2} & \textbf{74.5} & 0.85 & 74.5 & \textbf{85.0} & 0.88 \\
    \midrule
    LAD-Drive~\cite{schmidt2026laddrive}    & IROS'26  & 83.5 & 87.0 & 0.95 & 71.3 & 78.1 & 0.89 & 49.8 & 58.3 & 0.86 & 68.2 & 74.5 & 0.90 \\
    \rowcolor{gray!15}
     \textbf{+ \emph{Expand-and-Prune} (Ours)}         &  & \textbf{85.4} & 88.3 & \textbf{0.96} & \textbf{83.4} & 87.4 & \textbf{0.94} & 58.9 & 67.7 & \textbf{0.89} & \textbf{75.9} & 81.1 & \textbf{0.93} \\
    \bottomrule
  \end{tabular}
\end{table*}

\begin{table*}[t]
\centering
\caption{
Component ablation on LangAuto. The selected configuration (gray) of each stage is propagated to the subsequent stage. 
}
\label{tab:component_ablation}
\footnotesize

\begin{subtable}[t]{0.34\textwidth}
\centering
\caption{Intent-guided candidate expansion.}
\label{tab:ablation_intent}

\setlength{\tabcolsep}{6.0pt}
\begin{tabular}{cc|ccc}
    \toprule
    Native & Intent & \multicolumn{3}{c}{Mean} \\
    Cand. & Cand. &
    DS $\uparrow$ & RC $\uparrow$ & IS $\uparrow$ \\
    \cmidrule(r){1-2}
    \cmidrule(l){3-5}

    20 & 0  & 68.2 & 74.5 & 0.90 \\
    \midrule
    20 & 10 & 71.1 & 77.1 & 0.91 \\
    20 & 20 & 69.6 & 75.4 & 0.91 \\
    \rowcolor{gray!15}
    20 & \textbf{30} & \textbf{72.1} & \textbf{79.5} & 0.89 \\
    20 & 40 & 71.3 & 77.1 & \textbf{0.92} \\
    20 & 50 & 71.2 & 78.0 & 0.91 \\
    0  & 30 & 62.8 & 69.9 & 0.86 \\
    0  & 50 & 64.4 & 73.3 & 0.86 \\
    \bottomrule
\end{tabular}
\end{subtable}
\hfill
\begin{subtable}[t]{0.31\textwidth}
\centering
\caption{Temporal-memory selection.}
\label{tab:ablation_temporal}

\setlength{\tabcolsep}{5.0pt}
\begin{tabular}{c|ccc}
    \toprule
    Memory & \multicolumn{3}{c}{Mean} \\
    Selection &
    DS $\uparrow$ & RC $\uparrow$ & IS $\uparrow$ \\
    \cmidrule(r){1-1}
    \cmidrule(l){2-4}

    None & 72.1 & 79.5 & 0.89 \\
    \midrule
    Top-1 & 72.7 & 76.9 & \textbf{0.94} \\
    Top-3 & 73.0 & 79.8 & 0.91 \\
    Top-5 & 72.1 & 74.1 & 0.93 \\
    Entropy $(1,3)$ & 70.9 & 77.1 & 0.91 \\
    \rowcolor{gray!15}
    \textbf{Entropy $(1,5)$} & \textbf{73.4} & \textbf{80.9} & 0.90 \\
    Top-$p$ ($p=.8$) & 71.3 & 78.1 & 0.91 \\
    Top-$p$ ($p=.9$) & 71.2 & 79.6 & 0.89 \\
    \bottomrule
\end{tabular}
\end{subtable}
\hfill
\begin{subtable}[t]{0.31\textwidth}
\centering
\caption{Adaptive candidate pruning.}
\label{tab:ablation_pruning}

\setlength{\tabcolsep}{5.0pt}
\begin{tabular}{c|ccc}
    \toprule
    Pruning & \multicolumn{3}{c}{Mean} \\
    Strategy &
    DS $\uparrow$ & RC $\uparrow$ & IS $\uparrow$ \\
    \cmidrule(r){1-1}
    \cmidrule(l){2-4}

    None & 73.4 & 80.9 & 0.90 \\
    \midrule
    Top-5  & 73.9 & 79.7 & 0.92 \\
    Top-10 & 73.1 & 77.8 & 0.92 \\
    Entropy $(3,10)$ & 75.2 & 79.2 & 0.92 \\
    \rowcolor{gray!15}
    \textbf{Entropy $(4,12)$} & \textbf{75.9} & \textbf{81.1} & \textbf{0.93} \\
    Entropy $(5,15)$ & 74.1 & 79.8 & 0.90 \\
    Top-$p$ ($p=.8$) & 73.7 & 79.2 & 0.92 \\
    Top-$p$ ($p=.9$) & 74.3 & 79.0 & \textbf{0.93} \\
    \bottomrule
\end{tabular}
\end{subtable}

\vspace{-0.75em}

\end{table*}

\noindent \textbf{Baselines and implementation.}
We evaluate \emph{Expand-and-Prune} on four state-of-the-art candidate-based planners: LAD-Drive~\cite{schmidt2026laddrive}, BridgeDrive~\cite{liu2026bridgedrive}, DriveSuprim~\cite{yao2026drivesuprim}, and SparseDriveV2~\cite{sun2026sparsedrivev2}, using their unmodified models as baselines. 
For the evaluated pretrained planners, baseline results are reproduced using the official checkpoints under our evaluation setup and hardware.
All adaptations are performed exclusively at test time, while the learned planner parameters and scoring functions remain unchanged.

Individual components are instantiated according to each planner's candidate representation and processing structure, as summarized in Tab.~\ref{tab:planner_adaptation}. 
Among the evaluated methods, LAD-Drive is the only planner that exposes an explicit probabilistic action-belief state, enabling direct evaluation of intent-guided candidate expansion. 
Moreover, its sparse trajectory anchors serve as initializations that are geometrically refined through diffusion, making both intent-guided and temporal candidates suitable additional hypotheses. 
BridgeDrive likewise operates on sparse trajectory anchors followed by diffusion-bridge refinement, for which we incorporate temporal memory before adaptive pruning.
In contrast, DriveSuprim and SparseDriveV2 follow a selection-based paradigm in which trajectories from dense, fixed candidate vocabularies are progressively scored and filtered rather than geometrically refined.
We therefore restrict their adaptation to pruning, since intent-guided or temporally propagated trajectories would largely duplicate geometry already represented in their native candidate sets.

\begin{figure*}[t]
    \centering
    \includegraphics[width=\textwidth]{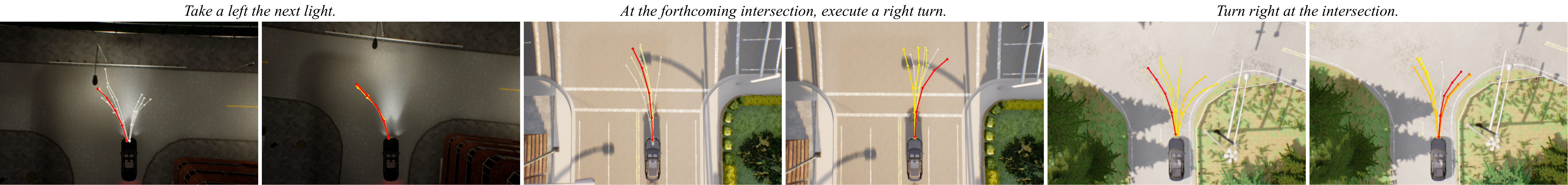}
    \caption{
    Qualitative comparison on LangAuto. Each pair compares LAD-Drive (left) with LAD-Drive + \emph{Expand-and-Prune} (right). Trajectory color encodes confidence from white (low) to red (high). The first scenario shows a guardrail collision during a left turn avoided through improved candidate coverage. In the latter two, route deviations are mitigated by better alignment with the planner's intent belief.
    }
    \label{fig:qualitative_results}
    
    \vspace{-0.5em}
    
\end{figure*}

\begin{table*}[t]
\centering
\caption{
Transfer of \emph{Expand-and-Prune} across planners and benchmarks. (a) BridgeDrive on Bench2Drive and Fail2Drive, where HM denotes the harmonic mean of Driving Score and Success Rate. (b) Adaptive pruning on DriveSuprim and SparseDriveV2 on NAVSIMv2.
}
\label{tab:transfer_results}
\footnotesize

\begin{subtable}[t]{0.70\textwidth}
    \centering
    \caption{Bench2Drive and Fail2Drive (F2D).}
    \label{tab:bridgedrive_transfer}

    \setlength{\tabcolsep}{6.0pt}
    \begin{tabular}{l cc ccc ccc}
        \toprule
        \multirow{2}{*}{Method}
        & \multicolumn{2}{c}{Bench2Drive}
        & \multicolumn{3}{c}{F2D Base}
        & \multicolumn{3}{c}{F2D Generalization} \\
        \cmidrule(lr){2-3}
        \cmidrule(lr){4-6}
        \cmidrule(l){7-9}

        & DS $\uparrow$
        & SR $\uparrow$
        & DS $\uparrow$
        & SR $\uparrow$
        & HM $\uparrow$
        & DS $\uparrow$
        & SR $\uparrow$
        & HM $\uparrow$ \\
        \midrule

        BridgeDrive
        & 95.9 & 88.2
        & 92.2 & 95.0 & 93.6
        & 80.2 & 73.0 & 76.4 \\

        + Pruning
        & 96.2 & 88.2
        & 92.0 & 95.0 & 93.5
        & 81.2 & 74.0 & 77.4 \\

        + Temporal
        & 96.2 & 88.6
        & 92.6 & 96.0 & 94.3
        & 82.0 & 75.0 & 78.3 \\

        \rowcolor{gray!15}
        \textbf{+ Temporal + Pruning}
        & \textbf{96.5} & \textbf{88.7}
        & \textbf{93.2} & \textbf{98.0} & \textbf{95.5}
        & \textbf{83.1} & \textbf{78.0} & \textbf{80.5} \\

        \bottomrule
    \end{tabular}
\end{subtable}
\hfill
\begin{subtable}[t]{0.27\textwidth}
    \centering
    \caption{NAVSIMv2.}
    \label{tab:navsim_transfer}

    \setlength{\tabcolsep}{6.0pt}
    \begin{tabular}{l c}
        \toprule
        Planner & EPDMS $\uparrow$ \\
        \midrule

        DriveSuprim (R34) & 83.07 \\
        \rowcolor{gray!15}
        \textbf{+ Pruning} & \textbf{83.67} \\

        \midrule

        DriveSuprim (ViT-L) & 87.03 \\
        \rowcolor{gray!15}
        \textbf{+ Pruning} & \textbf{87.27} \\

        \midrule

        SparseDriveV2 & 87.04 \\
        \rowcolor{gray!15}
        \textbf{+ Pruning} & \textbf{87.37} \\

        \bottomrule
    \end{tabular}
\end{subtable}

\vspace{-1.25em}

\end{table*}

\subsection{Candidate-Space Adaptation on LAD-Drive}
\label{sec:laddrive_results}

\noindent \textbf{Main results.}
Tab.~\ref{tab:prompt_config} compares \emph{Expand-and-Prune} with LAD-Drive and existing approaches on LangAuto. 
Applying \emph{Expand-and-Prune} to the unchanged LAD-Drive planner increases the mean DS from $68.2$ to $75.9$, corresponding to an absolute improvement of $7.7$ points. The gain is accompanied by a substantial increase in RC from $74.5$ to $81.1$, while the IS simultaneously improves from $0.90$ to $0.93$. 
Thus, the higher route completion is not accompanied by a degradation in infraction score.

The improvement becomes particularly pronounced on longer routes. 
While DS improves by $1.9$ points on Tiny routes, \emph{Expand-and-Prune} yields improvements of $12.1$ and $9.1$ points on Short and Long routes, respectively. 
These gains are primarily accompanied by higher route completion, which increases by $9.3$ points on Short and $9.4$ points on Long routes, while IS also improves consistently across all route categories. 
The stronger gains on longer routes suggest that candidate-space adaptation is increasingly beneficial over extended closed-loop horizons, where maintaining suitable, temporally consistent trajectory hypotheses is crucial.

Overall, \emph{Expand-and-Prune} establishes a new state of the art on LangAuto in mean DS ($75.9$) and IS ($0.93$).
Notably, it also surpasses LMGenDrive~\cite{shao2026lmgendrive} ($74.5$ mean DS), which improves language-conditioned driving through a generative world model and a multi-stage training pipeline. 
In contrast, \emph{Expand-and-Prune} obtains these gains solely through test-time candidate-space adaptation without retraining or modifying the learned planner parameters. 

\noindent \textbf{Component ablations.}
As shown in Tab.~\ref{tab:ablation_intent}, augmenting LAD-Drive's $20$ native anchors with intent-guided candidates improves mean DS for all evaluated expansion budgets. 
The best configuration adds $30$ candidates, increasing DS from $68.2$ to $72.1$ and RC from $74.5$ to $79.5$. 
Further increasing the budget yields no additional gains, indicating that the improvement does not simply result from processing more hypotheses. 
Importantly, replacing the native anchors with $30$ or $50$ intent-guided candidates reduces DS to $62.8$ and $64.4$, respectively.
Hence, the two candidate sources are complementary: native anchors provide broad geometric coverage, while intent-guided expansion increases candidate density in semantically relevant regions of the motion space.

We next add temporal memory to the selected expansion configuration and compare fixed Top-$K$, entropy-adaptive, and Top-$p$ selection in Tab.~\ref{tab:ablation_temporal}.
Top-$p$ retains the smallest set of highest-scoring candidates whose cumulative normalized probability exceeds a threshold $p$.
Entropy adaptation with bounds $(1,5)$ achieves the best overall result, improving DS from $72.1$ to $73.4$ and RC from $79.5$ to $80.9$, while increasing IS from $0.89$ to $0.90$. 
Fixed budgets exhibit non-monotonic behavior: Top-1 and Top-3 improve DS, whereas Top-5 reduces both DS and RC, indicating that simply increasing temporal-memory capacity is not uniformly beneficial.
Overall, the selected entropy-based strategy outperforms both the evaluated fixed Top-$K$ and adaptive Top-$p$ variants, indicating that entropy is a more effective criterion for adapting temporal memory capacity in our setting.

Finally, Tab.~\ref{tab:ablation_pruning} evaluates pruning on top of intent-guided expansion and entropy-adaptive temporal memory. 
The selected entropy-based configuration $(4,12)$ improves DS from $73.4$ to $75.9$, while simultaneously increasing RC from $80.9$ to $81.1$ and IS from $0.90$ to $0.93$. 
Thus, pruning not only reduces the number of hypotheses forwarded to subsequent refinement, but can also improve planning quality. 
Fixed Top-$K$ pruning shows a less favorable trade-off: Top-5 slightly improves DS and IS while reducing RC, whereas Top-10 falls below the unpruned configuration in DS. 

Overall, the ablations reveal a complementary progression: intent-guided expansion broadens semantic support, temporal memory introduces scene-relevant hypotheses from the preceding frame, and uncertainty-aware pruning concentrates subsequent refinement on a dynamically sized subset of promising candidates, with representative baseline failures corrected by
\emph{Expand-and-Prune} illustrated in Fig.~\ref{fig:qualitative_results}.

\subsection{Transfer Across Benchmarks and Planners}
\label{sec:transfer}

\noindent \textbf{Bench2Drive and Fail2Drive benchmark.}
Tab.~\ref{tab:bridgedrive_transfer} evaluates the transfer of \emph{Expand-and-Prune} to BridgeDrive, including a component-wise analysis of temporal memory and adaptive pruning.
On the Bench2Drive benchmark, where the baseline already achieves a high DS of $95.9$, temporal memory and pruning provide modest individual gains, while their combination reaches the best result with $96.5$ DS and $88.7$ SR.
The effect is more pronounced on the Fail2Drive benchmark. 
Pruning alone remains close to the baseline on the base split but improves all metrics under generalization, whereas temporal memory yields stronger gains on both splits. 
Combining both components performs best throughout, increasing HM from $93.6$ to $95.5$ on the base split and from $76.4$ to $80.5$ under generalization, including gains of $+2.9$ DS and $+5.0$ SR in the latter setting.
These results reproduce the complementary behavior already observed on LAD-Drive: temporal memory enriches the candidate set with temporally relevant hypotheses, while pruning concentrates subsequent processing on a smaller set of promising candidates. 
The larger gains under Fail2Drive generalization further suggest that candidate-space adaptation becomes particularly valuable in out-of-distribution scenarios.

\noindent \textbf{NAVSIM benchmark.}
Tab.~\ref{tab:navsim_transfer} evaluates adaptive pruning on the selection-based DriveSuprim and SparseDriveV2 planners on NAVSIMv2.
Despite already employing fixed candidate-selection budgets, adaptive pruning consistently improves EPDMS across all evaluated models.
For DriveSuprim, performance increases from $83.07$ to $83.67$ with the ResNet-34 backbone and from $87.03$ to $87.27$ with ViT-L, indicating that the benefit is retained across different visual encoders. 
SparseDriveV2 similarly improves from $87.04$ to $87.37$ EPDMS.
These results complement the LAD-Drive and BridgeDrive experiments by showing that \emph{Expand-and-Prune} also transfers to dense, selection-based planners for which candidate expansion is not applied. 
Here, adapting only the downstream candidate capacity is sufficient to consistently improve planning quality. 

\subsection{Candidate-Space Analysis and Efficiency}
\label{sec:candidate_analysis}

\begin{figure}[t!]
\centering
    \includegraphics[width=1.0\linewidth]{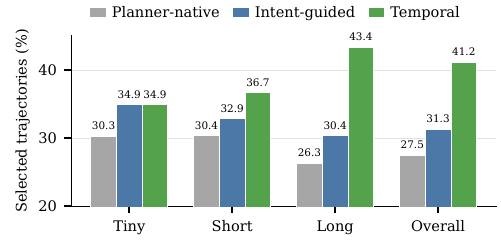}
    \vspace{-1.5em}
    \caption{Provenance of trajectories across route-length categories.} 
    \label{fig:candidate_provenance}

    \vspace{-1.05em}
    
\end{figure}

\noindent \textbf{Candidate provenance and adaptive budgets.}
We first analyze which candidate sources ultimately determine the selected trajectory. 
As shown in Fig.~\ref{fig:candidate_provenance}, only $27.5\%$ of selected trajectories originate from LAD-Drive's native geometric anchors, whereas $31.3\%$ originate from intent-guided candidates and $41.2\%$ from temporal memory. 
Hence, $72.5\%$ of final selections originate from hypotheses introduced by our candidate-space expansion. 
Temporal candidates are particularly effective: despite constituting only about $8.3\%$ of the expanded candidate set on average, they account for more than $40\%$ of selected trajectories. 
Their contribution also increases with route length, from $34.9\%$ on Tiny to $43.4\%$ on Long routes. 
This trend is consistent with the pronounced route-completion gains observed on longer routes in Tab.~\ref{tab:prompt_config}, suggesting that temporal hypotheses are particularly useful for maintaining consistent closed-loop behavior over extended horizons. \smallskip

We further analyze the candidate capacities selected by the entropy-adaptive
memory and pruning mechanisms. 
Averaged over all evaluation frames, temporal memory retains $4.6$ trajectories, while adaptive pruning forwards $8.1$ candidates to subsequent refinement.
To determine whether the gains merely result from identifying favorable global candidate budgets, we construct a matched fixed-capacity baseline using Top-5 temporal memory and Top-8 pruning, corresponding to the nearest integer values of these empirical averages.
The fixed configuration achieves $74.0$ DS, $80.2$ RC, and $0.90$ IS, compared with $75.9$ DS, $81.1$ RC, and $0.93$ IS for the adaptive configuration. 
Hence, a well-chosen fixed operating point already performs strongly, but adapting candidate capacity to frame-level uncertainty provides additional gains beyond the average budget alone.

\noindent \textbf{Runtime analysis.}
Tab.~\ref{tab:runtime_analysis} reports trajectory-head and total inference latency, separating the computational effect of candidate-space adaptation from architecture-specific backbone costs.
On LAD-Drive, intent-guided expansion and temporal memory increase head latency from $\SI{6.0}{\milli\second}$ to $\SI{9.2}{\milli\second}$ as additional hypotheses are processed. Adaptive pruning reduces the head latency to $\SI{8.2}{\milli\second}$ ($-11\%$), while simultaneously yielding the best planning performance. 
Consequently, the complete framework incurs only $\SI{2.2}{\milli\second}$ additional end-to-end latency over the original LAD-Drive baseline.
On BridgeDrive, temporal memory and adaptive pruning slightly reduce head latency from $\SI{73.4}{\milli\second}$ to $\SI{73.1}{\milli\second}$ and total latency from $\SI{95.3}{\milli\second}$ to $\SI{94.9}{\milli\second}$. Thus, the additional temporal hypotheses can be incorporated without increasing the overall inference cost when combined with adaptive pruning.
For the selection-based planners, adaptive pruning directly reduces the cost of their existing refinement stages. 
On DriveSuprim, refinement latency decreases from $\SI{2.0}{\milli\second}$ to $\SI{1.6}{\milli\second}$ with ResNet-34, while the smaller effect on total latency reflects the dominant cost of the visual backbone. 
SparseDriveV2 exhibits a more direct end-to-end benefit, reducing head latency from $\SI{19.0}{\milli\second}$ to $\SI{17.3}{\milli\second}$ and total latency from $\SI{23.8}{\milli\second}$ to $\SI{22.0}{\milli\second}$. 
Together with the performance improvements reported above, these results show that uncertainty-aware pruning can reduce candidate-processing cost beyond existing fixed candidate-selection budgets while improving planning quality.

\begin{table}[t!]
\centering
\caption{
Runtime on an NVIDIA L40S GPU. Head latency denotes the planner-specific
candidate-processing stages.
}
\label{tab:runtime_analysis}
\footnotesize

\begin{tabular}{l cc}
\toprule
Method & Head [ms] $\downarrow$ & Total [ms] $\downarrow$ \\
\midrule
LAD-Drive & \textbf{6.0} & \textbf{45.0} \\
 + Intent & 8.4 & 47.5 \\
 + Intent + Temporal & 9.2 & 48.5 \\
\rowcolor{gray!15}
 \textbf{+ Intent + Temporal + Pruning} & 8.2 & 47.2 \\
\midrule
BridgeDrive & 73.4 & 95.3 \\
\rowcolor{gray!15}
 \textbf{+ Temporal + Pruning} & \textbf{73.1} & \textbf{94.9} \\
\midrule
DriveSuprim (R34) & 2.0 & 88.7 \\
\rowcolor{gray!15}
 \textbf{+ Pruning} & \textbf{1.6} & \textbf{88.4} \\
\midrule
SparseDriveV2 & 19.0 & 23.8 \\
\rowcolor{gray!15}
 \textbf{+ Pruning} & \textbf{17.3} & \textbf{22.0} \\
\bottomrule
\end{tabular}

\vspace{-0.75em}

\end{table}

\section{Conclusion}
\label{sec:conclusion}

We presented \emph{Expand-and-Prune}, a test-time framework for adapting the candidate space of pretrained end-to-end driving planners without updating their learned parameters. 
The framework addresses two complementary dimensions of candidate-based planning: expansion adapts \emph{where} motion hypotheses are represented using semantic intent and temporal memory, while uncertainty-aware pruning adapts \emph{how many} candidates receive further computation. 
Across LAD-Drive, BridgeDrive, DriveSuprim, and SparseDriveV2, our experiments demonstrate that test-time candidate-space adaptation transfers across sparse generative and dense selection-based planners, improving planning quality while adaptive pruning also reduces candidate-processing cost.
Candidate-provenance and matched-capacity analyses further show that the gains arise not merely from processing more hypotheses or selecting a favorable fixed budget, but from introducing complementary candidates and adapting their capacity to the current prediction uncertainty. 
These results establish test-time candidate-space adaptation as a complementary direction for improving pretrained end-to-end driving planners without architecture-specific retraining.

\noindent \textbf{Limitations.}
Intent-guided expansion is currently evaluated only on LAD-Drive, which provides an explicit probabilistic action belief, and temporal memory is restricted to the immediately preceding frame. 
Future work will extend intent guidance to navigation commands or continuous goals and explore longer-term context through trajectory histories or planner-native latent representations.

\noindent \textbf{Acknowledgments.} This work has been funded by the BMFTR under support code 13FH544KB2 (ADRIVE-GPT). The authors thank the DACHS data analysis cluster, hosted at Esslingen University and co-funded by the MWK within the DFG's "Großgeräte der Länder" program, for providing the computational resources necessary for this research.

{
    \small
    \bibliographystyle{ieeenat_fullname}
    \bibliography{main}

@String(PAMI = {IEEE Trans. Pattern Anal. Mach. Intell.})

@String(CVPR= {IEEE Conf. Comput. Vis. Pattern Recog.})

@String(ICCV= {Int. Conf. Comput. Vis.})

@String(ECCV= {Eur. Conf. Comput. Vis.})

@String(NIPS= {Adv. Neural Inform. Process. Syst.})

@String(ICLR = {Int. Conf. Learn. Represent.})

@String(AAAI = {AAAI})

@String(CoRL = {Conf. Robot Learn.})

@String(ICRA = {IEEE Int. Conf. Robot. Autom.})

@String(IROS = {IEEE Int. Conf. Intell. Robots Syst.})

@String(PAMI  = {IEEE TPAMI})

@String(CVPR  = {CVPR})

@String(ICCV  = {ICCV})

@String(ECCV  = {ECCV})

@String(NIPS  = {NeurIPS})

@String(ICLR  = {ICLR})

@String(CoRL = {CoRL})

@String(ICRA = {ICRA})

@String(IROS = {IROS})

@inproceedings{shao2024lmdrive,
    title={Lmdrive: Closed-loop end-to-end driving with large language models},
    author={Shao, Hao and Hu, Yuxuan and Wang, Letian and Song, Guanglu and Waslander, Steven L and Liu, Yu and Li, Hongsheng},
    booktitle=CVPR,
    pages={15120--15130},
    year=2024
}

@inproceedings{winter2025bevdriver,
    title={BEVDriver: Leveraging BEV Maps in LLMs for Robust Closed-Loop Driving},
    author={Winter, Katharina and Azer, Mark and Flohr, Fabian B},
    booktitle=IROS,
    year=2025,
}

@article{jia2024bench2drive,
    title={Bench2drive: Towards multi-ability benchmarking of closed-loop end-to-end autonomous driving},
    author={Jia, Xiaosong and Yang, Zhenjie and Li, Qifeng and Zhang, Zhiyuan and Yan, Junchi},
    journal=NIPS,
    volume=37,
    pages={819--844},
    year=2024
}

@inproceedings{dosovitskiy2017carla,
    title = { {CARLA}: {An} Open Urban Driving Simulator},
    author = {Alexey Dosovitskiy and German Ros and Felipe Codevilla and Antonio Lopez and Vladlen Koltun},
    booktitle = CoRL,
    pages = {1--16},
    year = 2017
}

@inproceedings{zhang2025vldrive,
  title={VLDrive: Vision-Augmented Lightweight MLLMs for Efficient Language-grounded Autonomous Driving},
  author={Zhang, Ruifei and Zhang, Wei and Tan, Xiao and Yang, Sibei and Wan, Xiang and Luo, Xiaonan and Li, Guanbin},
  booktitle=ICCV,
  pages={5923--5933},
  year=2025
}

@inproceedings{zhang2025adadrive,
  title={AdaDrive: Self-Adaptive Slow-Fast System for Language-Grounded Autonomous Driving},
  author={Zhang, Ruifei and Xie, Junlin and Zhang, Wei and Chen, Weikai and Tan, Xiao and Wan, Xiang and Li, Guanbin},
  booktitle=ICCV,
  pages={5112--5121},
  year=2025
}

@article{dauner2024navsim,
  title={Navsim: Data-driven non-reactive autonomous vehicle simulation and benchmarking},
  author={Dauner, Daniel and Hallgarten, Marcel and Li, Tianyu and Weng, Xinshuo and Huang, Zhiyu and Yang, Zetong and Li, Hongyang and Gilitschenski, Igor and Ivanovic, Boris and Pavone, Marco and others},
  journal=NIPS,
  volume=37,
  pages={28706--28719},
  year=2024
}

@inproceedings{schmidt2026laddrive,
    title={LAD-Drive: Bridging Language and Trajectory with Action-Aware Diffusion Transformers}, 
    author={Schmidt, Fabian and Fedurko, Karol and Enzweiler, Markus and Valada, Abhinav},
    booktitle=IROS,
    year=2026,
}

@inproceedings{yao2026drivesuprim,
  title={Drivesuprim: Towards precise trajectory selection for end-to-end planning},
  author={Yao, Wenhao and Li, Zhenxin and Lan, Shiyi and Wang, Zi and Sun, Xinglong and Alvarez, Jose M and Wu, Zuxuan},
  booktitle=AAAI,
  volume=40,
  number=14,
  pages={11910--11918},
  year=2026}

@inproceedings{sun2026sparsedrivev2,
  title={Sparsedrivev2: Scoring is all you need for end-to-end autonomous driving},
  author={Sun, Wenchao and Lin, Xuewu and Chen, Keyu and Pei, Zixiang and Li, Xiang and Shi, Yining and Zheng, Sifa},
  booktitle=ECCV,
  year=2026
}

@article{li2025gtrs,
  title={Generalized trajectory scoring for end-to-end multimodal planning},
  author={Li, Zhenxin and Yao, Wenhao and Wang, Zi and Sun, Xinglong and Chen, Joshua and Chang, Nadine and Shen, Maying and Wu, Zuxuan and Lan, Shiyi and Alvarez, Jose M},
  journal={arXiv preprint arXiv:2506.06664},
  year={2025}
}

@inproceedings{liao2025diffusiondrive,
  title={Diffusiondrive: Truncated diffusion model for end-to-end autonomous driving},
  author={Liao, Bencheng and Chen, Shaoyu and Yin, Haoran and Jiang, Bo and Wang, Cheng and Yan, Sixu and Zhang, Xinbang and Li, Xiangyu and Zhang, Ying and Zhang, Qian and others},
  booktitle=CVPR,
  pages={12037--12047},
  year=2025,
  organization={IEEE}
}

@article{zou2025diffusiondrivev2,
  title={Diffusiondrivev2: Reinforcement learning-constrained truncated diffusion modeling in end-to-end autonomous driving},
  author={Zou, Jialv and Chen, Shaoyu and Liao, Bencheng and Zheng, Zhiyu and Song, Yuehao and Zhang, Lefei and Zhang, Qian and Liu, Wenyu and Wang, Xinggang},
  journal={arXiv preprint arXiv:2512.07745},
  year={2025}
}

@inproceedings{xing2025goalflow,
  title={Goalflow: Goal-driven flow matching for multimodal trajectories generation in end-to-end autonomous driving},
  author={Xing, Zebin and Zhang, Xingyu and Hu, Yang and Jiang, Bo and He, Tong and Zhang, Qian and Long, Xiaoxiao and Yin, Wei},
  booktitle=CVPR,
  pages={1602--1611},
  year=2025,
  organization={IEEE}
}

@inproceedings{liu2026bridgedrive,
  title={Bridgedrive: Diffusion bridge policy for closed-loop trajectory planning in autonomous driving},
  author={Liu, Shu and Chen, Wenlin and Li, Weihao and Wang, Zheng and Yang, Lijin and Huang, Jianing and Huang, Zhongzhan and Cheng, Ze and Yang, Hao},
  booktitle=ICLR,
  volume=2026,
  pages={130989--131009},
  year=2026
}

@article{sima2025centaur,
  title={Centaur: Robust end-to-end autonomous driving with test-time training},
  author={Sima, Chonghao and Chitta, Kashyap and Yu, Zhiding and Lan, Shiyi and Luo, Ping and Geiger, Andreas and Li, Hongyang and Alvarez, Jose M},
  journal={arXiv preprint arXiv:2503.11650},
  year={2025}
}

@article{xu2026toad,
  title={Test-Time Trajectory Optimization for Autonomous Driving},
  author={Xu, Yihong and Zablocki, Eloi and Yin, Yuan and Ramzi, Elias and Kirby, Ellington and Boulch, Alexandre and Cord, Matthieu},
  journal={arXiv preprint arXiv:2606.07170},
  year={2026}
}

@article{yao2026had,
  title={HAD: Combining Hierarchical Diffusion with Metric-Decoupled RL for End-to-End Driving},
  author={Yao, Wenhao and Sun, Xinglong and Li, Zhenxin and Lan, Shiyi and Wang, Zi and Alvarez, Jose M and Wu, Zuxuan},
  journal={arXiv preprint arXiv:2604.03581},
  year={2026}
}

@inproceedings{kirby2026drivor,
  title={Driving on registers},
  author={Kirby, Ellington and Boulch, Alexandre and Xu, Yihong and Yin, Yuan and Puy, Gilles and Zablocki, {\'E}loi and Bursuc, Andrei and Gidaris, Spyros and Marlet, Renaud and Bartoccioni, Florent and others},
  booktitle=CVPR,
  pages={32058--32069},
  year=2026
}

@inproceedings{kim2026storm,
  title={SToRM: Supervised Token Reduction for Multi-modal LLMs toward efficient end-to-end autonomous driving},
  author={Kim, Seo Hyun and Park, Jin Bok and Koo, Do Yeon and Park, Ho Gun and Chun, Il Yong},
  booktitle=ICRA,
  year=2026
}

@article{zhang2024adh,
  title={Ad-h: Autonomous driving with hierarchical agents},
  author={Zhang, Zaibin and Tang, Shiyu and Zhang, Yuanhang and Fu, Talas and Wang, Yifan and Liu, Yang and Wang, Dong and Shao, Jing and Wang, Lijun and Lu, Huchuan},
  journal={arXiv preprint arXiv:2406.03474},
  year={2024}
}

@article{you2026samoe,
  title={SAMoE-VLA: A Scene Adaptive Mixture-of-Experts Vision-Language-Action Model for Autonomous Driving},
  author={You, Zihan and Liu, Hongwei and Dang, Chenxu and Wang, Zhe and Ang, Sining and Wang, Aoqi and Wang, Yan},
  journal={arXiv preprint arXiv:2603.08113},
  year={2026}
}

@article{liu2025dsdrive,
  title={Dsdrive: Distilling large language model for lightweight end-to-end autonomous driving with unified reasoning and planning},
  author={Liu, Wenru and Liu, Pei and Ma, Jun},
  journal={arXiv preprint arXiv:2505.05360},
  year={2025}
}

@article{shao2026lmgendrive,
  title={LMGenDrive: Bridging Multimodal Understanding and Generative World Modeling for End-to-End Driving},
  author={Shao, Hao and Wang, Letian and Zhou, Yang and Hu, Yuxuan and Zong, Zhuofan and Waslander, Steven L and Zhan, Wei and Li, Hongsheng},
  journal={arXiv preprint arXiv:2604.08719},
  year={2026}
}

@article{gerstenecker2026fail2drive,
  title={Fail2drive: Benchmarking closed-loop driving generalization},
  author={Gerstenecker, Simon and Geiger, Andreas and Renz, Katrin},
  journal={arXiv preprint arXiv:2604.08535},
  year={2026}
}

@inproceedings{Cao2025navsimv2, 
	author = {Wei Cao and Marcel Hallgarten and Tianyu Li and Daniel Dauner and Xunjiang Gu and Caojun Wang and Yakov Miron and Marco Aiello and Hongyang Li and Igor Gilitschenski and Boris Ivanovic and Marco Pavone and Andreas Geiger and Kashyap Chitta}, 
	title = {Pseudo-Simulation for Autonomous Driving}, 
	booktitle = CoRL, 
	year = 2025, 
}

@article{chitta2022transfuser,
  title={Transfuser: Imitation with transformer-based sensor fusion for autonomous driving},
  author={Chitta, Kashyap and Prakash, Aditya and Jaeger, Bernhard and Yu, Zehao and Renz, Katrin and Geiger, Andreas},
  journal=PAMI,
  volume=45,
  number=11,
  pages={12878--12895},
  year=2022,
  publisher={IEEE}
}

@article{wu2022tcp,
  title={Trajectory-guided control prediction for end-to-end autonomous driving: A simple yet strong baseline},
  author={Wu, Penghao and Jia, Xiaosong and Chen, Li and Yan, Junchi and Li, Hongyang and Qiao, Yu},
  journal=NIPS,
  volume=35,
  pages={6119--6132},
  year=2022
}

@inproceedings{yin2026diffrefiner,
  title={Diffrefiner: Coarse to fine trajectory planning via diffusion refinement with semantic interaction for end to end autonomous driving},
  author={Yin, Liuhan and Ju, Runkun and Guo, Guodong and Cheng, Erkang},
  booktitle=AAAI,
  volume=40,
  number=14,
  pages={12009--12017},
  year=2026
}

@inproceedings{yang2024diffusiones,
  title={Diffusion-es: Gradient-free planning with diffusion for autonomous and instruction-guided driving},
  author={Yang, Brian and Su, Huangyuan and Gkanatsios, Nikolaos and Ke, Tsung-Wei and Jain, Ayush and Schneider, Jeff and Fragkiadaki, Katerina},
  booktitle=CVPR,
  pages={15342--15353},
  year=2024,
  organization={IEEE}
}

@inproceedings{hu2023uniad,
  title={Planning-oriented autonomous driving},
  author={Hu, Yihan and Yang, Jiazhi and Chen, Li and Li, Keyu and Sima, Chonghao and Zhu, Xizhou and Chai, Siqi and Du, Senyao and Lin, Tianwei and Wang, Wenhai and others},
  booktitle=CVPR,
  pages={17853--17862},
  year=2023,
  organization={IEEE}
}

@article{li2024hydramdp,
  title={Hydra-mdp: End-to-end multimodal planning with multi-target hydra-distillation},
  author={Li, Zhenxin and Li, Kailin and Wang, Shihao and Lan, Shiyi and Yu, Zhiding and Ji, Yishen and Li, Zhiqi and Zhu, Ziyue and Kautz, Jan and Wu, Zuxuan and others},
  journal={arXiv preprint arXiv:2406.06978},
  year=2024
}

@inproceedings{li2025hydranext,
  title={Hydra-next: Robust closed-loop driving with open-loop training},
  author={Li, Zhenxin and Wang, Shihao and Lan, Shiyi and Yu, Zhiding and Wu, Zuxuan and Alvarez, Jose M},
  booktitle=ICCV,
  pages={27305--27314},
  year=2025,
  organization={IEEE}
}

@inproceedings{vitelli2022safetynet,
  title={Safetynet: Safe planning for real-world self-driving vehicles using machine-learned policies},
  author={Vitelli, Matt and Chang, Yan and Ye, Yawei and Ferreira, Ana and Wo{\l}czyk, Maciej and Osi{\'n}ski, B{\l}a{\.z}ej and Niendorf, Moritz and Grimmett, Hugo and Huang, Qiangui and Jain, Ashesh and others},
  booktitle=ICRA,
  pages={897--904},
  year=2022,
  organization={IEEE}
}

@article{guo2025ipad,
  title={ipad: Iterative proposal-centric end-to-end autonomous driving},
  author={Guo, Ke and Liu, Haochen and Wu, Xiaojun and Pan, Jia and Lv, Chen},
  journal={arXiv preprint arXiv:2505.15111},
  year={2025}
}

@inproceedings{feng2026rap,
  title={Rap: 3d rasterization augmented end-to-end planning},
  author={Feng, Lan and Gao, Yang and Zablocki, Eloi and Li, Quanyi and Li, Wuyang and Liu, Sichao and Cord, Matthieu and Alahi, Alexandre},
  booktitle=ICLR,
  volume=2026,
  pages={37722--37737},
  year=2026
}

@article{song2026diver,
  title={Diver: Reinforced diffusion breaks imitation bottlenecks in end-to-end autonomous driving},
  author={Song, Ziying and Liu, Lin and Pan, Hongyu and Liao, Bencheng and Guo, Mingzhe and Yang, Lei and Zhang, Yongchang and Xu, Shaoqing and Jia, Caiyan and Luo, Yadan},
  journal=PAMI,
  year=2026,
  publisher={IEEE}
}

@article{li2025ztrs,
  title={Ztrs: Zero-imitation end-to-end autonomous driving with trajectory scoring},
  author={Li, Zhenxin and Yao, Wenhao and Wang, Zi and Sun, Xinglong and Chen, Jingde and Chang, Nadine and Shen, Maying and Song, Jingyu and Wu, Zuxuan and Lan, Shiyi and others},
  journal={arXiv preprint arXiv:2510.24108},
  year={2025}
}

@inproceedings{jiang2026vadv2,
  title={VADv2: End-to-end vectorized autonomous driving via probabilistic planning},
  author={Jiang, Bo and Chen, Shaoyu and Gao, Hao and Liao, Bencheng and Zhang, Qian and Liu, Wenyu and Wang, Xinggang},
  booktitle=ICLR,
  volume=2026,
  pages={68886--68900},
  year=2026
}

@inproceedings{jiang2023vad,
  title={Vad: Vectorized scene representation for efficient autonomous driving},
  author={Jiang, Bo and Chen, Shaoyu and Xu, Qing and Liao, Bencheng and Chen, Jiajie and Zhou, Helong and Zhang, Qian and Liu, Wenyu and Huang, Chang and Wang, Xinggang},
  booktitle=ICCV,
  pages={8306--8316},
  year=2023,
  organization={IEEE}
}

@article{schmidt2025graphpilot,
  title={Graphpilot: Grounded scene graph conditioning for language-based autonomous driving},
  author={Schmidt, Fabian and Enzweiler, Markus and Valada, Abhinav},
  journal={arXiv preprint arXiv:2511.11266},
  year={2025}
}
}

\clearpage
\maketitlesupplementary

This supplementary material provides additional details and analyses for
\emph{Expand-and-Prune}. We first describe the planner-specific integration
and implementation settings for LAD-Drive~\cite{schmidt2026laddrive},
BridgeDrive~\cite{liu2026bridgedrive}, DriveSuprim~\cite{yao2026drivesuprim},
and SparseDriveV2~\cite{sun2026sparsedrivev2}. We then provide additional
details on the evaluation benchmarks and metrics, followed by extended
component ablations for LAD-Drive and a detailed NAVSIMv2 metric breakdown
for DriveSuprim and SparseDriveV2. Finally, we present additional qualitative
examples from the LangAuto closed-loop evaluation.

\section{Implementation Details}

All adaptations are applied exclusively at inference time using the pretrained model weights and original scoring functions. 
For each entropy-adaptive component, the candidate capacity is determined from the normalized entropy of the corresponding score distribution as described in the main paper. 
Below, we detail the planner-specific integration and selected capacity bounds. 
The LAD-Drive configuration is selected through the extended ablations reported in Sec.~\ref{sec:supp_ablations}.

\noindent \textbf{LAD-Drive.}
LAD-Drive \cite{schmidt2026laddrive} initializes its diffusion planner from $20$ trajectory anchors and refines them through two denoising steps. 
We augment the native candidate set with $30$ intent-guided candidates allocated across the action-specific banks according to the probabilistic meta-action belief. 
In addition, temporal memory propagates high-confidence trajectories from the preceding planning step after transforming them into the current ego coordinate frame.
The number of temporal candidates is selected entropy-adaptively within $K_{\mathrm{mem}}\in[1,5]$. Adaptive pruning is applied after the first of the two denoising steps, where the intermediate trajectory scores are used to retain $K_{\mathrm{prune}}\in[4,12]$ candidates according to their normalized entropy. 
Only the retained trajectories are processed by the second denoising step.

\noindent \textbf{BridgeDrive.}
BridgeDrive \cite{liu2026bridgedrive} employs an anchor-guided diffusion-bridge policy that iteratively transforms coarse trajectory anchors into scene-conditioned plans over $20$ diffusion timesteps. 
We augment its anchor-based hypothesis set with motion-compensated trajectories propagated from the preceding planning step. 
The temporal-memory capacity is selected entropy-adaptively within $K_{\mathrm{mem}}\in[1,3]$.
After the first diffusion-bridge update, we apply adaptive pruning to the intermediate candidate scores and retain $K_{\mathrm{prune}}\in[3,10]$ trajectories. 
The remaining diffusion updates are then performed only on the retained hypotheses. 
We do not apply intent-guided expansion to BridgeDrive because it does not expose an explicit probabilistic semantic-intent belief analogous to LAD-Drive.

\noindent \textbf{DriveSuprim.}
DriveSuprim \cite{yao2026drivesuprim} follows a coarse-to-fine trajectory-selection architecture.
Its Trajectory Decoder first scores the complete trajectory vocabulary, after which the original model forwards the fixed top-$256$ candidates to a Refinement Decoder for fine-grained rescoring. 
We replace this fixed selection with entropy-adaptive pruning directly after coarse scoring and before the Refinement Decoder. 
Based on the entropy of the coarse trajectory-score distribution, we retain $K_{\mathrm{prune}}\in[16,64]$ candidates. 
The trajectory vocabulary, coarse scoring stage, and subsequent Refinement Decoder remain unchanged.

\noindent \textbf{SparseDriveV2.}
SparseDriveV2 \cite{sun2026sparsedrivev2} factorizes its trajectory vocabulary into geometric paths and velocity profiles and performs progressive coarse scoring independently over the two branches before composing the retained elements into trajectory hypotheses. 
We replace its fixed Top-$K$ operations with entropy-adaptive selection at each filtering stage, computing the capacity independently from the corresponding path or velocity score distribution. For the path branch, the first filtering stage retains $K_{\mathrm{path}}^{(0)}\in[4,128]$ candidates instead of a fixed Top-$128$, while the subsequent stage uses $K_{\mathrm{path}}^{(1)}\in[4,20]$ instead of Top-$20$. 
For the velocity branch, the first stage uses $K_{\mathrm{vel}}^{(0)}\in[12,64]$ instead of Top-$64$, and the final velocity filtering stage uses $K_{\mathrm{vel}}^{(1)}\in[4,15]$ instead of the fixed Top-$10$ used for NAVSIMv2. 
The retained path and velocity candidates are subsequently composed and processed by SparseDriveV2's original fine-grained trajectory-scoring stage.

\section{Benchmark and Metric Details}
\label{sec:supp_benchmarks}

We provide additional details on the four evaluation benchmarks and
their corresponding metrics used in the main paper.

\noindent \textbf{LangAuto.}
LangAuto~\cite{shao2024lmdrive} evaluates language-conditioned driving
in closed-loop CARLA simulation across all eight publicly available towns
and diverse environmental conditions. We use the standard LangAuto track
and report results separately for three route-length categories: Tiny routes
are shorter than $150\,\mathrm{m}$, Short routes range from $150$ to
$500\,\mathrm{m}$, and Long routes exceed $500\,\mathrm{m}$.
Performance is measured using Route Completion (RC), Infraction Score
(IS), and Driving Score (DS). RC denotes the percentage of the prescribed
route completed by the agent, while IS penalizes collisions and traffic-rule
violations through multiplicative penalty factors. Driving Score combines
progress and safety as
\begin{equation}
    \mathrm{DS}
    =
    \mathrm{RC}\cdot\mathrm{IS},
\end{equation}
where $\mathrm{RC}\in[0,100]$ and $\mathrm{IS}\in[0,1]$, such that
$\mathrm{DS}\in[0,100]$. We use DS as the primary LangAuto metric and
additionally report RC and IS to distinguish changes in route progress
from changes in driving safety.

\noindent \textbf{Bench2Drive.}
Bench2Drive~\cite{jia2024bench2drive} evaluates closed-loop driving
performance across $44$ interactive scenario types. Each scenario is
instantiated using five short routes of approximately $150\,\mathrm{m}$,
resulting in $N_{\mathrm{routes}}=220$ evaluation routes in total.

We report Success Rate (SR) and Driving Score (DS). A route is considered
successful if the agent reaches its destination within the allotted time
without incurring a traffic-rule infraction. Let
$N_{\mathrm{success}}$ denote the number of successfully completed routes
and $N_{\mathrm{routes}}$ the total number of evaluated routes. Since SR
is reported as a percentage, it is given by
\begin{equation}
    \mathrm{SR}
    =
    100\,
    \frac{N_{\mathrm{success}}}
         {N_{\mathrm{routes}}}.
\end{equation}
Thus, SR measures the percentage of evaluation routes completed without
failure or traffic-rule violations.

Driving Score additionally accounts for partial route completion and
infractions. For each route $i$, let
$\mathrm{RC}_i\in[0,100]$ denote the percentage of the route completed,
$N_i^{\mathrm{inf}}$ the number of infractions incurred on that route,
and $p_{i,j}\in(0,1]$ the penalty factor associated with the $j$-th
infraction. The Driving Score is computed as
\begin{equation}
    \mathrm{DS}
    =
    \frac{1}{N_{\mathrm{routes}}}
    \sum_{i=1}^{N_{\mathrm{routes}}}
    \mathrm{RC}_i
    \prod_{j=1}^{N_i^{\mathrm{inf}}}
    p_{i,j}.
\end{equation}
For routes without infractions, the penalty product evaluates to one.
Consequently, DS jointly reflects driving progress and traffic-rule
compliance, while SR provides a stricter measure of fully successful
route completion.

\noindent \textbf{Fail2Drive.}
Fail2Drive~\cite{gerstenecker2026fail2drive} evaluates closed-loop
generalization under controlled distribution shifts. The benchmark
contains $200$ routes in CARLA Town~13, organized into $100$ paired
in-distribution and generalization routes. Within each pair, road geometry,
spawn locations, and traffic goals remain fixed, while the generalization
route introduces a targeted distribution shift. The benchmark comprises
$17$ novel scenario classes grouped into robustness, visual generalization
for lateral control, visual generalization for longitudinal control, and
behavioral generalization. We refer to the in-distribution evaluation as
\emph{Base} in the main paper and report both Base and Generalization
performance.

Following the Bench2Drive protocol, Fail2Drive reports DS and SR.
Route Completion is not reported separately because the routes are
intentionally short and nearly always finishable. In addition, Fail2Drive
reports the harmonic mean (HM) of DS and SR,
\begin{equation}
    \mathrm{HM}
    =
    \frac{
        2\,\mathrm{DS}\,\mathrm{SR}
    }{
        \mathrm{DS}+\mathrm{SR}
    },
\end{equation}
where DS and SR are expressed on the same $0$--$100$ scale. HM therefore
provides a balanced summary of driving quality and successful route
completion under both in-distribution and shifted conditions.

\noindent \textbf{NAVSIMv2.}
NAVSIMv2 \cite{Cao2025navsimv2} evaluates trajectory planning on the real-world NAVSIM
\emph{navtest} split using the Extended Predictive Driver Model Score
(EPDMS). Compared with the original NAVSIM evaluation, NAVSIMv2 extends
the set of planning metrics with Driving-Direction Compliance (DDC),
Traffic-Light Compliance (TLC), Lane Keeping (LK), and Extended Comfort
(EC), while the original comfort metric is replaced by History Comfort
(HC). Together with No-At-Fault Collision (NC), Drivable-Area Compliance
(DAC), Ego Progress (EP), and Time to Collision (TTC), these metrics are
aggregated as
\begin{equation}
\begin{aligned}
    \mathrm{EPDMS}
    ={}&
    \mathrm{NC}\cdot
    \mathrm{DAC}\cdot
    \mathrm{DDC}\cdot
    \mathrm{TLC}
    \\
    &\cdot
    \frac{
        5(\mathrm{EP}+\mathrm{TTC})
        +2(\mathrm{LK}+\mathrm{HC}+\mathrm{EC})
    }{16}.
\end{aligned}
\end{equation}
The four compliance metrics therefore act as multiplicative penalties,
whereas progress, time to collision, lane keeping, and comfort are combined
through a weighted average. The resulting normalized score lies in
$[0,1]$; following standard benchmark reporting, we report EPDMS after
scaling by $100$.

\section{Extended Results}

\subsection{LAD-Drive Component Ablations}
\label{sec:supp_ablations}

We provide extended ablations for the three components of \emph{Expand-and-Prune} on LAD-Drive. 
Following the sequential evaluation protocol used in the main paper, we first determine the configuration for intent-guided candidate expansion, then evaluate temporal memory on top of the selected expansion setting, and finally evaluate adaptive pruning on top of both selected components. 
The selected configuration of each stage is therefore carried forward to the subsequent ablation.

\begin{table*}[t]
\centering
\caption{
Extended ablation of intent-guided candidate expansion on LangAuto. 
\emph{Prob.} denotes probabilistic allocation according to the action belief, while \emph{WTA} uses only the most probable action.
Entropy $(a,b)$ denotes an adaptive intent budget bounded by $a$ and $b$.
Gray denotes the selected configuration; best mean results are bold.
}
\label{tab:supp_intent_ablation}
\footnotesize

\begin{tabular}{lcc ccc ccc ccc | ccc}
\toprule
& \multicolumn{2}{c}{Candidates}
& \multicolumn{3}{c}{Tiny}
& \multicolumn{3}{c}{Short}
& \multicolumn{3}{c}{Long}
& \multicolumn{3}{|c}{Mean} \\
\cmidrule(lr){2-3}
\cmidrule(lr){4-6}
\cmidrule(lr){7-9}
\cmidrule(lr){10-12}
\cmidrule(lr){13-15}
Allocation
& Native & Intent Budget
& DS $\uparrow$ & RC $\uparrow$ & IS $\uparrow$
& DS $\uparrow$ & RC $\uparrow$ & IS $\uparrow$
& DS $\uparrow$ & RC $\uparrow$ & IS $\uparrow$
& DS $\uparrow$ & RC $\uparrow$ & IS $\uparrow$ \\
\midrule

-- & 20 & 0
& 83.50 & 87.00 & 0.95
& 71.30 & 78.10 & 0.89
& 49.80 & 58.30 & 0.86
& 68.20 & 74.47 & 0.90 \\

\midrule

\multirow{7}{*}{Prob.}
& 20 & 10
& 84.92 & 87.63 & 0.96
& 74.89 & 81.06 & 0.90
& 53.43 & 62.56 & 0.86
& 71.08 & 77.08 & 0.91 \\

& 20 & 20
& 81.20 & 82.06 & 0.98
& 70.16 & 77.63 & 0.89
& 57.37 & 66.38 & 0.87
& 69.58 & 75.36 & 0.91 \\

& \cellcolor{gray!15}20 & \cellcolor{gray!15}30
& \cellcolor{gray!15}81.09 & \cellcolor{gray!15}85.17 & \cellcolor{gray!15}0.95
& \cellcolor{gray!15}76.82 & \cellcolor{gray!15}83.14 & \cellcolor{gray!15}0.88
& \cellcolor{gray!15}58.51 & \cellcolor{gray!15}70.18 & \cellcolor{gray!15}0.83
& \cellcolor{gray!15}\textbf{72.14} & \cellcolor{gray!15}\textbf{79.50} & \cellcolor{gray!15} 0.89 \\

& 20 & 40
& 78.83 & 82.16 & 0.96
& 74.79 & 80.41 & 0.93
& 60.23 & 68.73 & 0.88
& 71.28 & 77.10 & \textbf{0.92} \\

& 20 & 50
& 83.30 & 88.49 & 0.95
& 73.85 & 81.42 & 0.89
& 56.38 & 64.07 & 0.89
& 71.17 & 77.99 & 0.91 \\

& 0 & 30
& 85.83 & 90.11 & 0.94
& 62.23 & 69.88 & 0.84
& 40.21 & 49.62 & 0.80
& 62.76 & 69.87 & 0.86 \\

& 0 & 50
& 83.19 & 87.14 & 0.95
& 63.12 & 73.76 & 0.82
& 46.74 & 58.95 & 0.80
& 64.35 & 73.28 & 0.86 \\

\midrule

WTA & 20 & 30
& 79.95 & 82.19 & 0.96
& 71.90 & 78.88 & 0.89
& 60.83 & 68.68 & 0.89
& 70.89 & 76.58 & 0.91 \\

\midrule

\multirow{3}{*}{Prob.}
& 20 & Entropy $(5,30)$
& 81.80 & 84.14 & 0.95
& 71.41 & 78.39 & 0.90
& 54.73 & 61.26 & 0.87
& 69.32 & 74.60 & 0.91 \\

& 20 & Entropy $(10,40)$
& 79.63 & 80.37 & 0.99
& 68.93 & 77.20 & 0.88
& 55.50 & 64.31 & 0.87
& 68.02 & 73.96 & 0.91 \\

& 20 & Entropy $(15,50)$
& 80.60 & 82.95 & 0.97
& 70.49 & 79.19 & 0.86
& 55.86 & 65.43 & 0.86
& 68.99 & 75.86 & 0.90 \\

\bottomrule
\end{tabular}
\end{table*}

\begin{table*}[t]
\centering
\caption{
Extended ablation of temporal-memory selection on LangAuto.
All variants use the selected intent-guided expansion with $20$ native and $30$ intent-guided candidates.
For entropy-based selection, $(K_{\min},K_{\max})$ denotes the adaptive memory bounds and $\tau=0.5$.
Gray denotes the selected configuration; best mean results are bold.
}
\label{tab:supp_temporal_ablation}
\footnotesize

\begin{tabular}{ll ccc ccc ccc | ccc}
\toprule
\multicolumn{2}{c}{Temporal Memory}
& \multicolumn{3}{c}{Tiny}
& \multicolumn{3}{c}{Short}
& \multicolumn{3}{c}{Long}
& \multicolumn{3}{|c}{Mean} \\
\cmidrule(lr){1-2}
\cmidrule(lr){3-5}
\cmidrule(lr){6-8}
\cmidrule(lr){9-11}
\cmidrule(lr){12-14}

Strategy & Setting
& DS $\uparrow$ & RC $\uparrow$ & IS $\uparrow$
& DS $\uparrow$ & RC $\uparrow$ & IS $\uparrow$
& DS $\uparrow$ & RC $\uparrow$ & IS $\uparrow$
& DS $\uparrow$ & RC $\uparrow$ & IS $\uparrow$ \\
\midrule

None & --
& 81.09 & 85.17 & 0.95
& 76.82 & 83.14 & 0.88
& 58.51 & 70.18 & 0.83
& 72.14 & 79.50 & 0.89 \\

\midrule

\multirow{3}{*}{Top-$K$}
& $1$
& 82.34 & 83.83 & 0.98
& 77.41 & 81.30 & 0.94
& 58.27 & 65.61 & 0.89
& 72.67 & 76.91 & \textbf{0.94} \\

& $3$
& 83.98 & 86.95 & 0.96
& 75.46 & 83.99 & 0.91
& 59.58 & 68.52 & 0.87
& 73.01 & 79.82 & 0.91 \\

& $5$
& 85.93 & 79.58 & 0.95
& 73.37 & 77.11 & 0.95
& 56.98 & 65.70 & 0.88
& 72.09 & 74.13 & 0.93 \\

\midrule

\multirow{2}{*}{Entropy}
& $(1,3)$
& 83.12 & 85.89 & 0.97
& 69.96 & 77.81 & 0.89
& 59.68 & 67.46 & 0.88
& 70.92 & 77.05 & 0.91 \\

& \cellcolor{gray!15}$(1,5)$
& \cellcolor{gray!15}82.40 & \cellcolor{gray!15}87.16 & \cellcolor{gray!15}0.94
& \cellcolor{gray!15}75.85 & \cellcolor{gray!15}84.61 & \cellcolor{gray!15}0.89
& \cellcolor{gray!15}61.90 & \cellcolor{gray!15}71.05 & \cellcolor{gray!15}0.88
& \cellcolor{gray!15}\textbf{73.39} & \cellcolor{gray!15}\textbf{80.94} & \cellcolor{gray!15}0.90 \\

\midrule

\multirow{2}{*}{Top-$p$}
& $0.8$
& 80.63 & 83.78 & 0.96
& 75.66 & 80.83 & 0.93
& 57.60 & 69.72 & 0.84
& 71.30 & 78.11 & 0.91 \\

& $0.9$
& 82.31 & 88.33 & 0.92
& 73.64 & 82.99 & 0.87
& 57.53 & 67.36 & 0.86
& 71.16 & 79.56 & 0.89 \\

\bottomrule
\end{tabular}


\end{table*}

\noindent \textbf{Intent-guided candidate expansion.}
Tab.~\ref{tab:supp_intent_ablation} provides the complete evaluation of intent-guided expansion. Starting from LAD-Drive's $20$ native anchors, probabilistic intent allocation improves mean DS for all evaluated fixed expansion budgets. Adding $30$ intent-guided candidates performs best, increasing mean DS from $68.20$ to $72.14$ and RC from $74.47$ to $79.50$.
Increasing the expansion budget further does not yield additional gains.
Importantly, removing the native anchors and relying exclusively on $30$ or $50$ intent-guided candidates substantially reduces performance, confirming that the two candidate sources are complementary rather than interchangeable.
Probabilistic allocation also outperforms winner-takes-all allocation at the same budget ($72.14$ vs.\ $70.89$ DS), supporting the use of the full meta-action belief instead of collapsing it to its most likely action.
Finally, adapting the total intent budget from the action-belief entropy does not improve over the fixed budget of $30$ candidates. 
We therefore retain $30$ probabilistically allocated intent-guided candidates for all subsequent experiments.

\begin{table*}[t]
\centering
\caption{
Extended ablation of adaptive pruning on LangAuto.
All variants use $20$ native and $30$ intent-guided candidates.
Unless stated otherwise, temporal memory uses entropy-adaptive selection with bounds $(1,5)$. Entropy-based memory and pruning use $\tau=0.5$.
Gray denotes the selected configuration; best mean results are bold.
The final row reports the matched fixed-capacity configuration.
}
\label{tab:supp_pruning_ablation}
\footnotesize

\begin{tabular}{lll ccc ccc ccc | ccc}
\toprule
Memory
& \multicolumn{2}{c}{Pruning}
& \multicolumn{3}{c}{Tiny}
& \multicolumn{3}{c}{Short}
& \multicolumn{3}{c}{Long}
& \multicolumn{3}{|c}{Mean} \\
\cmidrule(lr){2-3}
\cmidrule(lr){4-6}
\cmidrule(lr){7-9}
\cmidrule(lr){10-12}
\cmidrule(lr){13-15}

& Strategy & Setting
& DS $\uparrow$ & RC $\uparrow$ & IS $\uparrow$
& DS $\uparrow$ & RC $\uparrow$ & IS $\uparrow$
& DS $\uparrow$ & RC $\uparrow$ & IS $\uparrow$
& DS $\uparrow$ & RC $\uparrow$ & IS $\uparrow$ \\
\midrule

Entropy $(1,5)$
& None & --
& 82.40 & 87.16 & 0.94
& 75.85 & 84.61 & 0.89
& 61.90 & 71.05 & 0.88
& 73.39 & 80.94 & 0.90 \\

\midrule

\multirow{12}{*}{Entropy $(1,5)$}

& \multirow{5}{*}{Top-$K$}
& $1$
& 65.25 & 69.51 & 0.91
& 63.86 & 71.64 & 0.89
& 40.66 & 48.32 & 0.80
& 56.59 & 63.16 & 0.87 \\

& & $3$
& 86.57 & 87.85 & 0.98
& 72.63 & 78.96 & 0.89
& 53.67 & 60.19 & 0.89
& 70.95 & 75.67 & 0.92 \\

& & $5$
& 83.20 & 85.57 & 0.97
& 80.97 & 87.91 & 0.89
& 57.60 & 65.66 & 0.88
& 73.92 & 79.72 & 0.92 \\

& & $10$
& 84.75 & 86.76 & 0.97
& 76.44 & 81.55 & 0.92
& 57.96 & 65.20 & 0.89
& 73.05 & 77.84 & 0.92 \\

& & $20$
& 82.75 & 87.25 & 0.95
& 78.99 & 82.35 & 0.92
& 58.04 & 68.95 & 0.84
& 73.26 & 79.51 & 0.90 \\

\cmidrule(lr){2-15}

& \multirow{4}{*}{Entropy}
& $(3,10)$
& 83.80 & 84.76 & 0.99
& 81.04 & 83.17 & 0.91
& 60.65 & 69.61 & 0.87
& 75.16 & 79.18 & 0.92 \\

& & \cellcolor{gray!15}$(4,12)$
& \cellcolor{gray!15}85.42 & \cellcolor{gray!15}88.25 & \cellcolor{gray!15}0.96
& \cellcolor{gray!15}83.39 & \cellcolor{gray!15}87.43 & \cellcolor{gray!15}0.94
& \cellcolor{gray!15}58.95 & \cellcolor{gray!15}67.72 & \cellcolor{gray!15}0.89
& \cellcolor{gray!15}\textbf{75.92} & \cellcolor{gray!15}\textbf{81.13} & \cellcolor{gray!15}\textbf{0.93} \\

& & $(5,15)$
& 83.20 & 87.15 & 0.95
& 81.01 & 86.63 & 0.89
& 58.14 & 65.72 & 0.87
& 74.11 & 79.83 & 0.90 \\

& & $(5,25)$
& 83.84 & 85.78 & 0.98
& 77.19 & 83.49 & 0.92
& 57.75 & 65.79 & 0.89
& 72.93 & 78.35 & \textbf{0.93} \\

\cmidrule(lr){2-15}

& \multirow{3}{*}{Top-$p$}
& $0.8$
& 83.92 & 85.73 & 0.97
& 75.78 & 83.11 & 0.90
& 61.27 & 68.67 & 0.88
& 73.66 & 79.17 & 0.92 \\

& & $0.9$
& 84.88 & 85.78 & 0.98
& 80.25 & 86.10 & 0.92
& 57.83 & 65.22 & 0.88
& 74.32 & 79.03 & \textbf{0.93} \\

& & $0.95$
& 78.97 & 80.25 & 0.98
& 76.02 & 81.50 & 0.93
& 59.14 & 68.97 & 0.84
& 71.38 & 76.91 & 0.92 \\

\midrule
\multicolumn{15}{l}{\textit{Matched fixed-capacity configuration}} \\

Top-$5$
& Top-$K$ & $8$
& 86.46 & 90.73 & 0.94
& 78.39 & 84.19 & 0.91
& 57.15 & 65.59 & 0.86
& 74.00 & 80.17 & 0.90 \\

\bottomrule
\end{tabular}
\end{table*}

\noindent \textbf{Temporal-memory selection.}
Using the selected $20+30$ native and intent-guided candidate configuration, Tab.~\ref{tab:supp_temporal_ablation} compares fixed Top-$K$ memory, entropy-adaptive memory, and Top-$p$ selection. 
Fixed Top-$1$ and Top-$3$ memory improve mean DS from $72.14$ to $72.67$ and $73.01$, respectively, while increasing the fixed memory size to five trajectories provides no further benefit. 
Entropy-adaptive memory with $(K_{\min},K_{\max})=(1,5)$ achieves the strongest overall result, reaching $73.39$ DS and $80.94$ RC. In comparison, the narrower entropy range $(1,3)$ and both evaluated Top-$p$ variants perform worse. 
Although Top-$1$ achieves the highest mean IS, its lower RC results in a lower overall DS.
We therefore use entropy-adaptive temporal memory with bounds $(1,5)$ in the remaining experiments.

\noindent \textbf{Adaptive pruning.}
Tab.~\ref{tab:supp_pruning_ablation} evaluates pruning on top of the selected intent-guided expansion and entropy-adaptive temporal memory. 
Fixed Top-$K$ pruning exhibits a clear sensitivity to the retained capacity: aggressive Top-$1$ pruning substantially degrades performance, while Top-$5$ provides a moderate improvement over the unpruned configuration. 
Entropy-adaptive pruning yields a more favorable trade-off. In particular, $(K_{\min},K_{\max})=(4,12)$ achieves the best mean performance with $75.92$ DS, $81.13$ RC, and $0.93$ IS. 
The alternative entropy ranges and Top-$p$ strategies remain below this configuration, showing that both the minimum retained capacity and the available adaptive capacity range are relevant.
To separate frame-level adaptation from the effect of selecting a favorable average candidate budget, we additionally evaluate a matched fixed-capacity configuration. 
The adaptive temporal memory retains $4.6$ trajectories on average and adaptive pruning forwards $8.1$ candidates in the full method.
We therefore compare against fixed Top-$5$ temporal memory followed by Top-$8$ pruning. 
This matched configuration reaches $74.00$ DS, compared with $75.92$ for the entropy-adaptive configuration, indicating that the gain cannot be explained solely by operating at a favorable average candidate capacity.

\subsection{DriveSuprim and SparseDriveV2 Extended Metrics}

Tab.~\ref{tab:supp_navsim_extended} provides the full NAVSIMv2 metric breakdown for DriveSuprim and SparseDriveV2. 
Across both DriveSuprim backbones and SparseDriveV2, adaptive pruning consistently improves EPDMS, further supporting the transferability of adaptive candidate selection across different selection-based planners. 
The improvements are accompanied by a consistent increase in several safety- and compliance-related metrics: no-at-fault collision, driveable-area compliance, and time to collision improve for all three configurations, while driving-direction and traffic-light compliance are maintained or slightly improved.

The gains are not uniform across all criteria. Ego progress and lane keeping decrease slightly for all three configurations, while history comfort remains unchanged. 
The largest EPDMS improvement is observed for DriveSuprim with ResNet-34, increasing from $83.069$ to $83.664$, while the ViT-L variant improves from $87.033$ to $87.264$ and SparseDriveV2 from $87.040$ to $87.368$. 
Overall, the metric breakdown indicates that adaptive pruning provides a favorable aggregate trade-off, with consistent improvements in several safety- and compliance-related criteria despite small reductions in progress and lane keeping.

\begin{table*}[t!]
\centering
\caption{
Extended NAVSIMv2 metric breakdown for DriveSuprim and SparseDriveV2.
NC denotes No-At-Fault Collision, DAC Driveable-Area Compliance, DDC Driving-Direction Compliance, TLC Traffic-Light Compliance, EP Ego Progress, TTC Time to Collision, LK Lane Keeping, HC History Comfort, and EC Two-Frame Extended Comfort.
Gray rows denote adaptive pruning. Best results within each baseline--pruning pair are highlighted in bold.
}
\label{tab:supp_navsim_extended}
\footnotesize

\begin{tabular}{l c | c c c c c c c c c}
\toprule
Method & EPDMS $\uparrow$ & NC $\uparrow$ & DAC $\uparrow$ & DDC $\uparrow$ & TLC $\uparrow$ & EP $\uparrow$ & TTC $\uparrow$ & LK $\uparrow$ & HC $\uparrow$ & EC $\uparrow$ \\
\midrule

DriveSuprim (ResNet-34) & 83.069 & 0.975 & 0.965 & 0.994 & 0.996 & \textbf{0.863} & 0.966 & \textbf{0.955} & \textbf{0.983} & \textbf{0.749} \\

\rowcolor{gray!15}
\quad + Adaptive Pruning & \textbf{83.664} & \textbf{0.981} & \textbf{0.975} & \textbf{0.995} & \textbf{0.997} & 0.861 & \textbf{0.973} & 0.952 & \textbf{0.983} & 0.743 \\

\midrule

DriveSuprim (ViT-L) & 87.033 & 0.984 & 0.986 & \textbf{0.996} & \textbf{0.998} & \textbf{0.904} & 0.978 & \textbf{0.970} & \textbf{0.983} & 0.766 \\

\rowcolor{gray!15}
\quad + Adaptive Pruning & \textbf{87.264} & \textbf{0.988} & \textbf{0.990} & \textbf{0.996} & \textbf{0.998} & 0.892 & \textbf{0.982} & 0.966 & \textbf{0.983} & \textbf{0.775} \\

\midrule

SparseDriveV2 & 87.040 & 0.985 & 0.982 & 0.996 & 0.998 & \textbf{0.904} & 0.979 & \textbf{0.972} & \textbf{0.983} & 0.781 \\

\rowcolor{gray!15}
\quad + Adaptive Pruning & \textbf{87.368} & \textbf{0.989} & \textbf{0.983} & \textbf{0.997} & \textbf{0.999} & 0.887 & \textbf{0.984} & 0.969 & \textbf{0.983} & \textbf{0.806} \\

\bottomrule
\end{tabular}
\end{table*}

\begin{figure*}[t!]
    \centering
    \includegraphics[width=\textwidth]{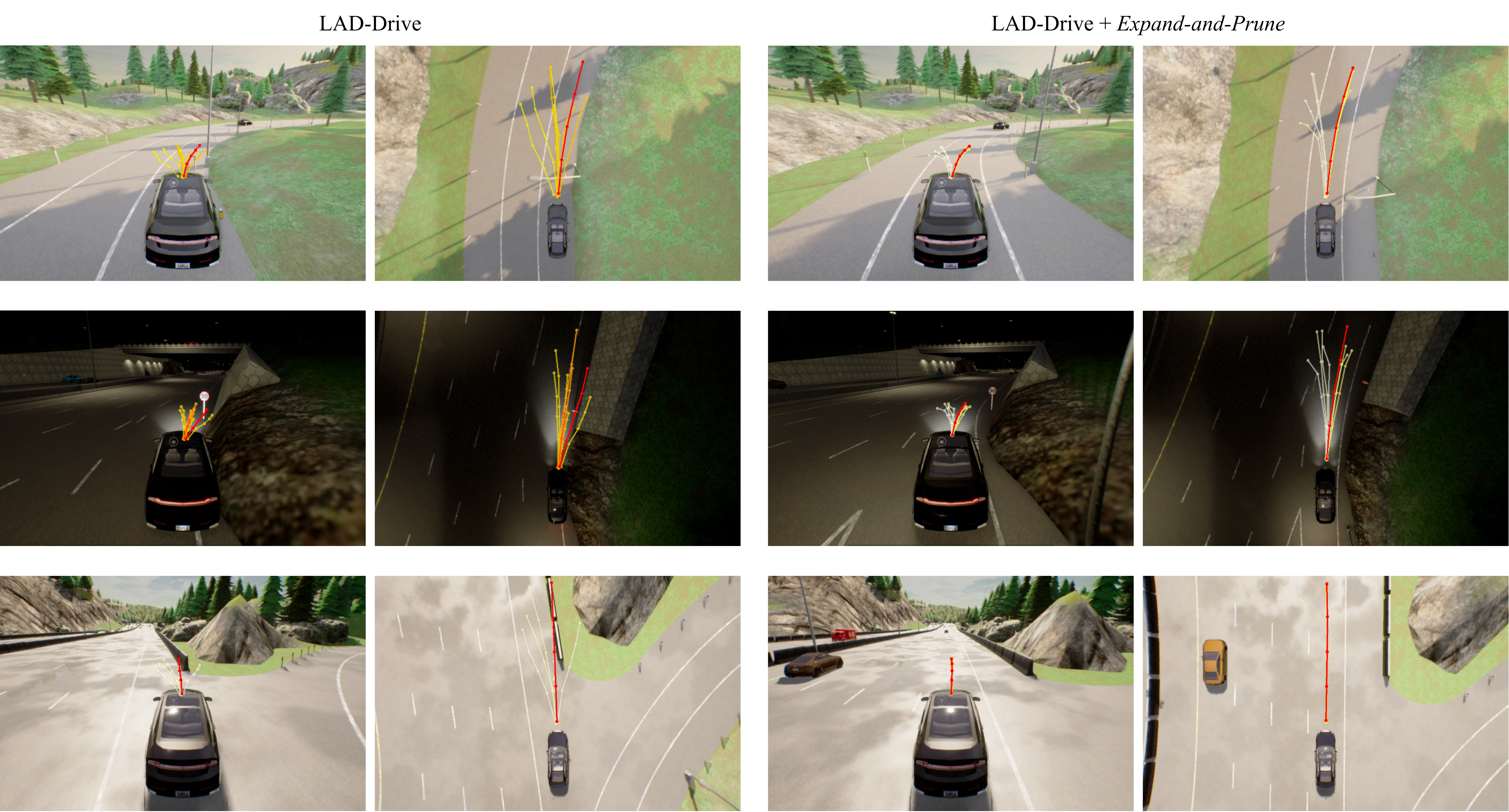}
    \caption{
    Additional qualitative comparisons on Long routes of LangAuto.
    For each scenario, the two left columns show LAD-Drive and the two
    right columns show LAD-Drive + \emph{Expand-and-Prune}, using a
    driving view and corresponding bird's-eye view.
    Trajectory color encodes confidence from white (low) to red (high).
    Top: in Scenario~36, LAD-Drive leaves the intended lane and collides
    with an object, while \emph{Expand-and-Prune} remains in the correct
    lane. Middle: in Scenario~37, LAD-Drive drifts out of its lane during
    a highway merge and collides with a traffic sign, whereas
    \emph{Expand-and-Prune} safely completes the merge. Bottom: in
    Scenario~43, ambiguity around a highway exit leads LAD-Drive into
    the guardrail, while \emph{Expand-and-Prune} stabilizes the prediction
    and continues straight.
    }
    \label{fig:qualitative_results_supp}
\end{figure*}

\section{Qualitative Results}

Fig.~\ref{fig:qualitative_results_supp} provides additional qualitative comparisons between LAD-Drive and LAD-Drive with \emph{Expand-and-Prune} on Long routes of LangAuto. 
For each scenario, the two left columns show LAD-Drive and the two right columns show \emph{Expand-and-Prune}, each using a driving view and the corresponding bird's-eye view. 
Candidate trajectories are visualized using a white-to-red confidence scale, where white denotes low confidence and red denotes high confidence.

In Scenario~36, LAD-Drive leaves the intended lane and continues along the road shoulder, eventually colliding with an object. 
In contrast, \emph{Expand-and-Prune} maintains a trajectory within the correct lane and safely traverses the same section. 
Scenario~37 illustrates a highway merge. LAD-Drive gradually drifts to the right, leaves its lane, and ultimately collides with a traffic sign. 
\emph{Expand-and-Prune} instead remains within the lane and follows the intended path, safely completing the merge.
Scenario~43 highlights ambiguity near a highway exit. 
LAD-Drive initially favors a right-exit maneuver before correcting its prediction toward continuing straight. 
However, the unstable trajectory prediction leads the vehicle into the guardrail. 
\emph{Expand-and-Prune} also represents the exit maneuver among its candidate hypotheses, but the additional intent-guided and temporal candidates provide stronger support for continuing straight, resulting in a stable trajectory that avoids the guardrail.

\end{document}